\documentclass{article}

    \PassOptionsToPackage{numbers, compress}{natbib}

\usepackage[preprint]{neurips_2026}

\usepackage[utf8]{inputenc} 
\usepackage[T1]{fontenc}    
\usepackage{hyperref}       
\usepackage{url}            
\usepackage{booktabs}       
\usepackage{amsfonts}       
\usepackage{nicefrac}       
\usepackage{microtype}      
\usepackage{xcolor}         

\usepackage{lineno,hyperref}
\usepackage[utf8]{inputenc} 
\usepackage{url}            
\usepackage{booktabs}       
\usepackage{amsfonts}       
\usepackage{nicefrac}       
\usepackage{microtype}      
\usepackage{longtable}      
\usepackage{graphicx}
\usepackage{amsmath}
\usepackage{booktabs}
\usepackage{algorithm}
\usepackage{algorithmic}
\usepackage{xcolor}
\definecolor{boston}{rgb}{0.8, 0.0, 0.0}
\usepackage{multirow}
\usepackage{amsfonts}
\usepackage{mathrsfs}  
\usepackage{bm}
\usepackage{setspace}
\usepackage{subcaption}
\usepackage{makecell}
\usepackage{mathtools}
\usepackage{graphicx}
\usepackage{xcolor}
\usepackage{tikz}
\usepackage{ragged2e}
\usepackage[normalem]{ulem}
\usepackage{makecell}

\title{ECG-NAT: A Self-supervised Neighborhood Attention Transformer for Multi-lead Electrocardiogram Classification}

\author{%
    Mahsa Gazeran$^{\dagger}$, 
  Sayvan Soleymanbaigi$^{\dagger}$, 
  Fatemeh Daneshfar$^{\dagger}$, \\
  \textbf{Amjad Seyedi}$^{\ddagger}$, 
  \textbf{Fardin Akhlaghian Tab}$^{\dagger, *}$ \\
  \\
  $^{\dagger}$Department of Computer Engineering, University of Kurdistan, Sanandaj, Iran \\
  $^{\ddagger}$Department of Mathematics and Operational Research, University of Mons, Mons, Belgium \\
  \texttt{\{mahsa.gazeran, s.soleymanbaigi, f.daneshfar, f.akhlaghian\}@uok.ac.ir} \\
  \texttt{seyedamjad.seyedi@umons.ac.be} \\
}

\begin{document}

\maketitle

\begin{abstract}
  Electrocardiogram (ECG) arrhythmia classification remains challenging due to signal variability, noise, limited labeled data, and the difficulty in achieving both accuracy and efficiency in models. While self-supervised learning reduces label dependency, most methods target either global contextual features or local morphological patterns, but rarely implement hierarchical multi-scale feature extraction. ECG signals require architectures that simultaneously capture fine-grained beat-level morphology and broader rhythm-level dependencies with computational efficiency. To overcome this limitation, this paper proposes the Electrocardiogram Neighborhood Attention Transformer (ECG-NAT), a novel self-supervised learning approach tailored for multi-lead ECG classification. Our two-stage approach begins with generative pretraining, using a masked autoencoder to reconstruct partially masked ECG signals across multiple diverse datasets, enabling the model to learn robust, domain-invariant representations from unlabeled data. This is followed by discriminative fine-tuning with a dual-loss function that combines supervised contrastive and cross-entropy losses, aligning representation learning with label prediction. The hierarchical attention mechanism efficiently captures multi-scale temporal features from localized beat morphology to broader rhythm patterns at low computational cost. ECG-NAT achieves robust performance on benchmark datasets, with 88.1\% accuracy using only 1\% labeled data, demonstrating strong efficacy in low-resource settings. The framework combines superior classification performance with computational efficiency, making it practical for real-time ECG diagnosis. The code will be made available upon acceptance at: \href{https://github.com/Mahsagazeran/ECG-NAT}{https://github.com/Mahsagazeran/ECG-NAT}.
\end{abstract}

\section{Introduction}
Cardiovascular disease (CVD) remains the foremost contributor to global mortality, responsible for an estimated 17.9 million lives lost each year, a staggering 32\% of all deaths worldwide~\cite{abubaker2022detection}. The electrocardiogram (ECG) serves as a vital diagnostic tool, enabling clinicians to detect and monitor cardiac abnormalities effectively~\cite{goy2013electrocardiography}. 
It is widely valued for its simplicity, cost efficiency, and non-invasive nature~\cite{paglini2012electrocardiography}. Numerous analysis methodologies have been devised to enhance the utility of ECG beyond its traditional applications. These include tasks for disease detection and classification \cite{dai2021convolutional}, annotation and localization \cite{han2023automated}, sleep staging ~\cite{rashidi2024strength}, biometric human identification~\cite{aleidan2023biometric}, and denoising~\cite{jin2024novel}, demonstrating the versatility and extensive application of ECG in modern medical diagnostics~\cite{luthra2019ecg}.

Accurate ECG labeling presents a significant challenge due to the complex, non-linear structure of cardiac signals, which is further complicated by the limited availability of annotated datasets that require time-consuming manual extraction and specialized clinical expertise \cite{fedjajevs2020platform}. As a solution, numerous machine learning models have been developed to assist in ECG interpretation. However, traditional approaches often fail to capture critical temporal patterns accurately and rely heavily on labeled data~\cite{minchole2019machine}. These limitations underscore the need for more efficient methods to leverage unlabeled data and decode intricate ECG waveform patterns \cite{zhai2018automated}.

To improve the capability of detecting complex data structures, deep learning techniques have been increasingly adopted. Deep learning has transformed this field by automating the extraction of meaningful patterns from complex data. Early approaches employed convolutional neural networks (CNNs)~\cite{zhai2018automated} to detect local signal features. For sequential data modeling, recurrent neural networks (RNNs) \cite{abrishami2018supervised} are naturally suited for ECG signals due to their recursive connections, which explicitly model temporal dependencies between consecutive heartbeats. Long short-term memory networks (LSTMs) \cite{saadatnejad2019lstm} extended this capability with gated mechanisms to better capture long-range rhythm patterns that standard RNNs often miss. More recently, Transformer-based approaches have demonstrated superior sequential modeling capabilities by processing complete input sequences in parallel through self-attention mechanisms. This paradigm shift enables more effective pattern recognition across all time steps, leading to improved performance in time-series classification tasks \cite{zhang2023token,cheng2023msw}.

Self-supervised learning (SSL) represents a significant advancement in machine learning, especially beneficial for fields with abundant unlabeled data \cite{qi2020small}. Specifically in ECG analysis, SSL techniques leverage unannotated datasets to derive meaningful representations that enhance signal classification and anomaly detection \cite{mehari2022self}.
A key technique in self-supervised learning, contrastive learning~\cite{wang2023adversarial}, strengthens feature representation by optimizing models to maximize the distance between dissimilar samples and minimize it for similar ones~\cite{chen2020simple}. Alternatively, generative approaches, such as Masked Autoencoders (MAE) \cite{hu2023spatiotemporal}, learn robust latent representations by reconstructing masked ECG segments. These self-supervised strategies enhance feature extraction and pattern discovery without requiring labeled data \cite{yang2022masked}.

Conventional transformer designs rely on global self-attention processes where each token interacts with all others in the sequence \cite{liang2021fusion}, 
 inherently resulting in $O(n^2)$ computational complexity \cite{parikh2016decomposable}, where $n$ is the number of tokens. For long sequences, this results in high computational and memory costs \cite{natarajan2020wide}. Local attention techniques address this by computing attention within fixed-size token windows centered at each position, achieving \(O(n k)\) complexity where \(k\) is the window size in tokens~\cite{dong2023arrhythmia}. Larger windows approach the computational cost of global attention, while smaller ones improve efficiency at the cost of potentially missing long-range dependencies.
Furthermore, recent techniques that integrate different forms of local attention have been devised. Some examples of transformers based on local attention are the Longformer \cite{beltagy2020longformer}, Swin Transformer \cite{liu2021swin}, and the newly launched Neighborhood Attention Transformer (NAT) \cite{hassani2023neighborhood}.  
 NAT employs fixed-window attention, where each token interacts only with nearby tokens in its neighborhood. This design was initially proven effective for image data, where local pixel relationships dominate. It builds hierarchical representations through successive layers, progressively capturing broader context without incurring the computational costs of global attention. 


While ECG data is inherently sequential, its periodicity exhibits recurring cyclical patterns with diagnostically important features spanning multiple temporal scales \cite{baek2019end}. Accurate arrhythmia classification requires simultaneously detecting localized beat abnormalities (e.g., ST-segment deviations, premature contractions) and sustained rhythm irregularities (e.g., atrial fibrillation) across multiple cardiac cycles a challenge that single-scale methods struggle to address. This makes hierarchical local attention mechanisms particularly suitable, as they can progressively expand their receptive field to capture both fine-grained morphology and broader rhythmic patterns while maintaining computational efficiency.
Most ECG classification methods critically depend on supervised learning, yet the large annotated datasets this requires are realistically unavailable~\cite{yang2024masked}. Self-supervised learning with local attention offers a practical solution by first training on widely available unlabeled ECG data and then fine-tuning with limited annotations. This strategy aligns with the inherent structure of the ECG and reduces the reliance on manually labeled examples.
\begin{figure}[]
  \centering
  \begin{subfigure}[b]{0.48\textwidth}
    \raisebox{1.5mm}{\includegraphics[width=\textwidth]{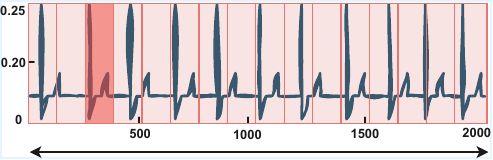}}
    \caption{Standard Self-Attention (SA)}
  \end{subfigure}
  \hspace{0.02\textwidth}
  \begin{subfigure}[b]{0.48\textwidth}
    \raisebox{0mm}{\includegraphics[width=\textwidth]{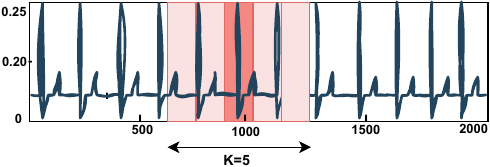}}
    \caption{Neighborhood Attention (NA)}
  \end{subfigure}
  \caption{\justifying Standard self-attention (SA) computes attention weights across all sequence positions, leading to quadratic complexity and potential loss of local cardiac patterns. Neighborhood Attention (NA), illustrated here with k=5 (k-nearest neighbors or k-NN), restricts computation to nearest neighbors only, reducing computational cost while better capturing the localized diagnostic features essential for ECG analysis.}
  \label{fig:enter-attention}
\end{figure}
Motivated by these insights, this work proposes ECG-NAT, a self-supervised framework that employs hierarchical local attention to learn multi-scale temporal representations, 
combined with a dual-loss fine-tuning strategy to optimize performance across diverse ECG classification tasks.


{The proposed design integrates a hierarchical local attention mechanism that substantially reduces computational overhead (illustrated in Figure~\ref{fig:enter-attention}) while capturing physiologically meaningful patterns such as QRS complexes and ST segments across multiple temporal scales. Our methodology is centered around three fundamental advances 
that function synergistically.
We first adapt a hierarchical local attention architecture tailored for multi-lead ECG signals, enabling multi-scale temporal feature extraction through progressive receptive field expansion, which is crucial for capturing both localized beat abnormalities and broader rhythm irregularities in arrhythmia classification. We substitute 2D patch embeddings with 1D convolutions in the tokenizer to process temporal ECG data, implement neighborhood attention across time steps in the NAT blocks, and utilize transposed convolutions in the decoder for signal reconstruction.  These improvements utilize NAT's local attention mechanism to efficiently collect morphological and rhythmic traits within physiologically pertinent neighborhoods, while substantially reducing processing requirements compared to global attention techniques. These improvements leverage localized attention to capture morphological and rhythm-related cues within physiologically relevant neighborhoods, while reducing computation compared to full self-attention.
Secondly, we utilize self-supervised pretraining with a masked autoencoder architecture, trained on various different unlabeled ECG datasets.  This multi-dataset methodology compels the model to acquire resilient, domain-invariant representations through the reconstruction of masked signal segments, facilitating generalization across diverse signal patterns without necessitating costly manual annotations.  Third, we implement a dual-loss optimization technique during fine-tuning that integrates supervised contrastive loss with cross-entropy loss.  This method concurrently improves inter-class feature distinction and promotes classification precision, allowing the model to optimize both discriminative representation learning and accurate label prediction.
 Collectively, these innovations establish a robust framework where the local attention mechanism tackles the inherent periodicity and computational limitations of the ECG signal, self-supervised pretraining mitigates the issue of label scarcity commonly found in clinical environments, and the dual-loss fine-tuning strategy converts generalized representations into task-specific classifiers. ECG-NAT achieves cutting-edge classification performance on challenging benchmarks such as PTB-XL~\cite{wagner2020ptb} and CPSC2018~\cite{liu2018open}, while simultaneously preserving computational efficiency and maintaining robust performance in low-resource clinical settings.}
To summarize, our key contributions are outlined below: 
\begin{itemize}
    \item ECG-NAT framework achieves computational efficiency and robust representation learning through hierarchical local attention that captures multi-scale temporal dependencies from beat-level morphology to rhythm-level patterns in multi-lead ECG signals, exploiting the inherent periodicity and hierarchical structure of cardiac activity.
    
    \item We propose a self-supervised framework combining local neighborhood attention and masked autoencoder reconstruction to capture localized ECG patterns like QRS complexes and ST segments, efficiently learning morphological features and generalizing across diverse datasets.

    \item Our fine-tuning stage employs a dual-loss strategy, combining contrastive and cross-entropy losses. This approach enhances the model's ability to differentiate among similar classes, thereby improving classification performance.

    \item By combining generative self-supervised learning with discriminative fine-tuning, ECG-NAT achieves state-of-the-art performance on challenging benchmarks like CPSC2018 and PTB-XL. The model also demonstrates strong efficacy in low-resource settings, delivering high performance even with very minimal labeled data.

\end{itemize}

This paper is organized into several sections. Section \ref{se:Related work} focuses on related work, while Section \ref{se:proposed model} provides a comprehensive description of the proposed approach and its architectural design. Section \ref{se:Experimental Setup} details the experimental configuration and presents findings. Finally, Section \ref{se:Conclusion} summarizes the work and suggests future avenues.

\section{Related Work}
\label{se:Related work}
This section discusses advanced ECG methods, including state-of-the-art transformer-based supervised models and recent self-supervised techniques, which enhance signal pattern capturing and data efficiency in ECG classification, thereby improving arrhythmia detection and classification.

\subsection{ECG analysis using supervised methods}
Under the supervised paradigm, numerous deep learning techniques have been proposed to classify ECG signals with labeled data.  Zhai \& Tin~\cite{zhai2018automated} proposed a method that utilizes the Continuous Wavelet Transform (CWT) in conjunction with CNN to transform ECG signals into time-frequency scalograms for arrhythmia classification. Meanwhile, Feyisa et al.~\cite{feyisa2022lightweight} developed a lightweight CNN with diverse receptive fields. This design, which incorporates filters of varied sizes and dilation rates, was developed to effectively model temporal features in cardiac signals. To address CNNs’ limitations in modeling long-range dependencies, recurrent models like RNNs and LSTMs have been explored. Saadatnejad et al.~\cite{saadatnejad2019lstm} designed a lightweight LSTM-based model with wavelet pre-processing and MLP fusion, optimized for real-time ECG applications. Abrishami et al.~\cite{abrishami2018supervised} applied bidirectional LSTM-RNNs to ECG interval segmentation, achieving accurate diagnostics with local features.
More recently, transformers have emerged as the dominant architectures for sequence modeling, leveraging self-attention to model long-range dependencies that surpass those of recurrent models. Natarajan et al.~\cite{natarajan2020wide} proposed a Transformer model with a combined wide and deep structure, incorporating both handcrafted and learned features to enable simultaneous prediction and classification of various conditions from 12-lead ECG recordings. Hu et al.~\cite{hu2022transformer} introduced a Transformer-CNN hybrid, framing arrhythmia detection as an object detection problem, which eliminates manual segmentation while supporting wearable use. El-Ghaish \& Eldele~\cite{el2024ecgtransform} further advanced this approach via ECGTransForm, which integrates multi-scale convolutions and bidirectional transformers.
Recent techniques for ECG classification increasingly use Vision Transformers (ViTs) because they effectively model global signal dependencies and capture diagnostically important local features. Dong et al.~\cite{dong2023arrhythmia} advanced transformer applications through ViTs with deformable attention, which enables more adaptable ECG morphology modeling and sets a new benchmark for multi-lead ECG classification effectiveness. Zhang et al.~\cite{zhang2023token} presented a multi-scale network that integrates both CNN and Transformer components, featuring a two-branch Transformer module for classifying 12-lead ECGs. This network combines learnable CNN blocks for multi-scale feature extraction and multi-scale embedding layers (MSEL) for patch-level representation. In contrast, Cheng et al.~\cite{cheng2023msw} developed the MSW-Transformer, a Transformer network that employs a multi-scale window shifting approach, where attention slides across different scales to collect diverse aspects of ECG signals, effectively addressing the variable frequency components inherent in ECG signals.
Concluding this review of supervised methodologies, transformer-based models have achieved leading performance in ECG classification. They accomplish this by utilizing self-attention to capture both localized and global features, thereby enabling robust modeling of complex waveform patterns.

\subsection{ECG analysis using self-supervised methods}
Self-supervised learning has revolutionized pre-trained models by enabling the extraction of rich features from unlabeled data without human annotations~\cite{qi2020small}. In this paradigm, models generate their supervisory signals through pretext tasks, learning features suitable for task-specific adaptation in downstream problems, often outperforming conventional supervised methods. 
This approach has become especially valuable in ECG classification due to the large quantity of unlabeled recordings and limited expert annotations. 
The most recent developments in self-supervised ECG classification learning involve examining contrastive and generative approaches. This facilitates the extraction of robust, broadly applicable features from unlabeled electrocardiogram signals. Contrastive methods learn by distinguishing between similar and dissimilar samples by creating different views of the same ECG signal through augmentations~\cite{wang2023adversarial}. Key examples include SimCLR~\cite{chen2020simple}, which uses various data augmentations; MoCo v3 (Momentum Contrast Version 3) ~\cite{chen2021empirical}, employing momentum encoders to improve representation quality; and CMSC (Contrastive Multi-segment Coding) \cite{kiyasseh2021clocs}, designed for multi-lead ECG data.  Eldele et al.~\cite{eldele2023self} developed TS-TCC, a technique for learning representations from time-series data that utilizes contrasting across temporal and contextual elements, and applied cross-view prediction to acquire robust features.  Le et al.~\cite{le2023scl} proposed sCL-ST, which enhances multi-label classification by using label information to correctly identify positive and negative pairs, addressing limitations of random negative sampling in traditional contrastive techniques.

Generative self-supervised methods utilize encoder-decoder architectures, where encoders extract representations from partially masked inputs, and decoders reconstruct missing portions~\cite{he2022masked}. In ECG classification, Masked Autoencoders (MAEs) have proven effective by randomly removing ECG signal segments and training models to predict these missing components~\cite{hu2023spatiotemporal}. Unlike contrastive methods requiring carefully designed augmentations, these approaches learn directly from signal structure and temporal dependencies, enabling models to leverage unlabeled ECG data while preserving physiologically significant features. Yang et al.~\cite{yang2022masked} combined multi-scale CNNs with transformers for acquiring both localized features and global patterns from ECG signals. This complementary approach enables better reconstruction of heavily masked inputs and produces superior representations for downstream ECG classification tasks.  Zhang et al.~\cite{zhang2023self} proposed the Cross Reconstruction Transformer (CRT), reconstructing dropped patches across time and frequency domains to capture meaningful temporal patterns. Yang et al.~\cite{yang2024masked} introduced MassMIB, an ECG representation learning approach based on masked self-supervision, which leverages a Multi-view Information Bottleneck methodology. This methodology processes time and frequency domain views simultaneously, applying information bottleneck principles to filter redundant information.  Zhang et al.~\cite{zhang2022maefe} presented MaeFE, a family of masked autoencoders tailored for ECG signals, comprising variations such as MTAE (a time-domain masked autoencoder), MLAE (a lead-dimension masked autoencoder), and MLTAE (a lead and time-domain masked autoencoder), each with a unique masking strategy.  Na et al.~\cite{na2024guiding} presented ST-MEM (Spatio-Temporal Masked Electrocardiogram Modeling), which reconstructs masked segments across spatial and temporal dimensions, employing lead-wise shared decoders to capture relationships between ECG leads.
In summary, generative self-supervised learning represents an advancement in ECG analysis by modeling signal patterns directly, without the need for complex data transformations. More advanced self-supervised approaches that integrate both contrastive and generative methods could further advance automated ECG classification, particularly in scenarios with limited labeled data, making them especially promising.

\section{Proposed Method} 
\label{se:proposed model}
\begin{figure}[t]
    \centering
    \includegraphics[width=1\linewidth]{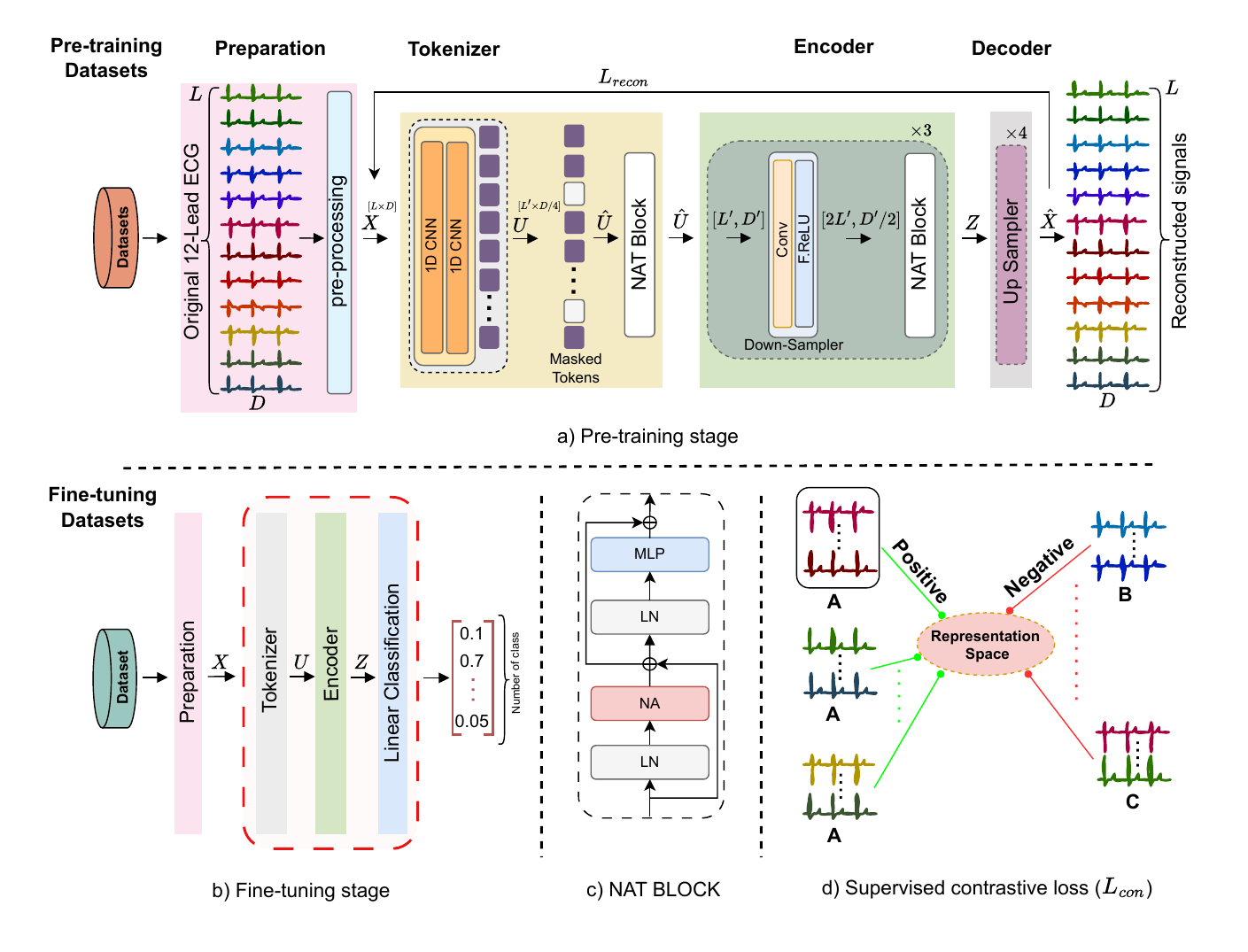}
    \caption{\justifying The ECG-NAT framework: a) The masked autoencoder architecture learns robust ECG representations by reconstructing masked signal segments, capturing meaningful patterns, and pretraining on standard unlabeled datasets enhances generalization capability; b) The pre-trained encoder is used for fine-tuning using a dual-loss approach, combining contrastive and cross-entropy losses; c) The NAT Block employs local attention mechanisms to extract contextually relevant features within neighborhoods; d) Contrastive loss pulls the source ECG signal closer to positive examples (A) from the same class while pushing it away from negative samples (B, C) of different classes, creating a representation space that effectively discriminates between ECG patterns. }
    \label{fig:enter-method}
\end{figure}

In this section, we present the proposed ECG-NAT framework, which is designed to efficiently learn and classify ECG signals through a combination of self-supervised representation learning and local attention mechanisms.
The ECG-NAT framework begins with thorough signal pre-processing to standardize and enhance the raw ECG signals. Following pre-processing, self-supervised pretraining creates robust representations. The subsequent fine-tuning stage includes supervised contrastive loss and linear classification for specific tasks. The local attention mechanism in Neighborhood Attention Transformer (NAT) effectively captures the critical localized waveform patterns in these pre-processed signals, which is a key component of accurate ECG classification. An overview of the ECG-NAT framework is shown in Figure~\ref{fig:enter-method}.
\subsection{Data Pre-processing}
The ECG pre-processing pipeline (Figure \ref{fig:enter-method}, preparation stage) ensures data quality before model input. Initially, a band-pass digital filter~\cite{constantinides1970spectral} with a frequency range of 0.5-40.0 Hz eliminates undesired noise; then, the signal amplitudes undergo normalization. The final step involves down-sampling from 500 Hz to 250 Hz, reducing the number of time points from 5000 to 2500. The standardized process is uniformly applied across all four datasets, ensuring a fair comparison with current best-performing baselines that follow analogous pre-processing protocols.
{To ensure uniform input dimensions across datasets with variable-length recordings, all ECG signals are standardized to 10 seconds (2500 time points). Shorter signals are zero-padded, while longer signals are segmented into non-overlapping 10-second windows, with the first segment used for classification.}
\subsection{Pre-training stage}
During pretraining, as depicted in Figure \ref{fig:enter-method}, we utilize a self-supervised autoencoder framework that leverages multiple unlabeled ECG datasets. This multi-dataset approach ensures our model generalizes well across varied signal patterns and prevents overfitting to characteristics specific to any single dataset. Following standard pre-processing, signal segments are randomly masked with Gaussian noise before input to the encoder. Subsequently, the decoder reconstructs the original signals from these corrupted inputs. This compels the model to acquire robust latent representations of essential ECG characteristics for downstream tasks.

\subsubsection{Encoder}
To get ECG signals ready for self-supervised learning, as illustrated in Figure~\ref{fig:enter-method} part~(a), we first tokenized the continuous input \( X 
\in \mathbb{R}^{L \times D} \), where \( L \) is the number of leads and \( D \) is the signal duration per lead. During tokenization, signals are segmented into overlapping windows. This results in a tokenized sequence \( U \in \mathbb{R}^{L' \times D'} \), where \( L' \) denotes the number of channels and \( D' \) indicates the spatial dimension.  The value $D'= D/4$ results from applying two consecutive CNN layers with a stride of 2, which reduces the token count by a factor of 4, as shown in the figure. Overlapping ensures continuity and preserves significant ECG properties across segment borders. A subset of these tokens is then randomly masked with Gaussian noise. Before being fed into the encoder, the masked tokens pass through a NAT block to further process the information. The encoder then receives these processed, noisy signals. Concurrently, the decoder undertakes reconstruction of the original, uncorrupted signal. Such strategic masking enables the model to acquire robust and meaningful representations that can generalize across diverse ECG patterns.
Building on this tokenization approach, the encoder establishes a hierarchical structure by utilizing localized attention mechanisms to efficiently process ECG data. The design interleaves attention blocks with CNN-based down-sampler modules, progressively reducing temporal resolution while expanding the receptive field to capture multi-scale temporal dependencies.

Figure~\ref{fig:enter-method}~(c) illustrates the NAT blocks, which begin by processing tokenized ECG signals. Here, each element undergoes three linear transformations, generating the query matrix $Q$, the key matrix $K$, and the value matrix $V$. Through this local attention mechanism, the model can effectively pinpoint and examine crucial patterns and relationships within ECG signals by focusing on the local vicinities of tokens. 
The attention weights for the $i$-th input, given a neighborhood size $k$, represented as $A^k_i$, are determined as:

\[
A^k_i =
\begin{bmatrix}
 Q_i K_{\mu_1(i)}^\top + B(i, \mu_1(i)) \\
 Q_i K_{\mu_2(i)}^\top + B(i, \mu_2(i)) \\
\vdots \\
 Q_i K_{\mu_k(i)}^\top + B(i, \mu_k(i))
\end{bmatrix},\tag{1}
\]
where $Q_i$ is the query vector associated with the $i$-th token, while $K_{\mu_j(i)}$ refers to the key vector for the $j$-th neighbor of token $i$. The notation $\mu_j(i)$ identifies the $j$-th nearest neighbor of token $i$. The term $B(i, \mu_j(i))$ incorporates relative positional biases between token $i$ and its neighbor $\mu_j(i)$, thereby capturing their spatial relationships during attention computation. The neighboring values matrix $V_i^k$ is built by gathering value vectors from token $i$'s $k$ nearest neighbors, where each row is a value projection of a neighboring token:

\[
V_i^k = 
\begin{bmatrix}
V_{\mu_1(i)}^\top & V_{\mu_2(i)}^\top & \cdots & V_{\mu_k(i)}^\top.
\end{bmatrix}^\top
\tag{2}
\]

Neighborhood Attention for the $i$-th token with neighborhood size $k$ is then defined as:

\[
NA_k(i) = \text{softmax} \left( \frac{A^k_i}{\sqrt{d}} \right) V^k_i,\tag{3}
\]
where $\sqrt{d}$ serves as the scaling parameter, utilized to normalize attention scores and maintain numerical stability during training. This process is applied to each ECG signal token.

Following the NAT blocks in the encoder, the down-sampler modules serve a key function in the hierarchical representation by reducing spatial dimensions while enriching feature channels, as shown in Figure~\ref{fig:enter-method} part(a). The down-sampler processes inputs with dimensions $[L', D']$, with $L'$ indicating the channel count and $D'$ representing the spatial dimension. Through convolutional layers (kernel-size 3, stride 2) and ReLU activation, these inputs are transformed to dimensions $[2L', D'/2]$, effectively halving the spatial size while doubling the channel count. This design integrates NAT blocks with CNN-based down-sampling to progressively reduce spatial resolution while expanding the receptive field, allowing the network to capture both local features and broader context.
The complete ECG-NAT architecture, through this encoder structure, accurately identifies key diagnostic patterns including abnormal waveforms, irregular rhythms, and segment distortions. This multi-scale analysis significantly improves arrhythmia classification across different types while preserving computational efficiency.
This hierarchical design progressively expands receptive fields through down-sampling, allowing ECG-NAT to maintain efficient local feature extraction while building broader temporal context crucial for capturing both beat-level morphology and rhythm-level patterns in ECG analysis.

\subsubsection{Decoder}
During pretraining, the decoder receives the latent representation \( Z \) from the encoder and reconstructs the original ECG signal \( \hat{X} \) through a sequence of transposed convolutional layers, as illustrated in Figure \ref{fig:enter-method} part(a). The reconstructed signal is then compared to the original input \( X \) using a mean squared error (MSE) loss, which encourages the model to learn robust feature representations that capture key temporal and morphological characteristics of the ECG. The MSE loss is formally defined as follows:

\[
L_{\text{recon}} = \sum_{i \in I} \| X_i - \hat{X_i} \|^2, \quad \hat{X} = \text{rec} \left( X \right),\tag{4}
\]
where the reconstruction loss \( L_{\text{recon}} \) quantifies the squared difference between the original signal values \( X_i \) and their reconstructed counterparts \( \hat{X}_i \) for each masked sample at index \( i \in I \). The set (I) comprises the index of the samples that were masked for reconstruction. 
This guides the model to recover the masked signal effectively. 
This reconstruction process enables the model to form robust representations of ECG patterns. The decoder's multi-layer CNN structure ensures smooth and accurate restoration of essential cardiac features, maintaining signal integrity even in the presence of noise.

\subsection{Fine-tuning with Supervised Contrastive Loss }
Task-specific fine-tuning investigates the ability of the pre-trained representations to transfer to downstream cardiac diagnostic tasks. In this stage, we investigate the extent to which features learned through self-supervised pretraining can be transferred to other classification goals. We use two complementary approaches: linear evaluation and full fine-tuning. For linear evaluation, the encoder, pre-trained in masked autoencoding, is frozen and used as a static feature extractor, while only the linear classification layer is trained. This process reveals the inherent characteristics of the representations learned without influencing them. Full fine-tuning, however, removes the restrictions on the pre-trained encoder and enables joint optimization of both the encoder and the classifier. Notably, the general representations achieved with pretraining inherently encompass vast amounts of morphological and rhythmic features with high discriminative strength. Fine-tuning significantly enhances these representations by tailoring them to the specifics of individual classification tasks, thereby refining decision boundaries and improving model performance. This composite ability, strong generalization via pretraining and specialized tuning via fine-tuning, illustrates the flexibility and efficacy of our self-supervised solution for electrocardiogram analysis.

{Unlike generative pretraining, this fine-tuning stage employs discriminative learning through supervised contrastive loss, which explicitly optimizes for class separation in the representation space.}
The fine-tuning methodology employs a joint training approach, where the encoder and classifier are optimized simultaneously using a combination of supervised contrastive loss and cross-entropy classification loss, as shown in Figure \ref{fig:enter-method}~(b). The encoder parameters evolve through backpropagation guided by both supervised contrastive loss and cross-entropy loss functions, creating a representation ${Z}$ that becomes increasingly tailored to specific ECG classification tasks. The supervised contrastive component plays a crucial role in this process by enhancing feature discrimination between different ECG classes. This optimization strategy is applied to multiple demanding ECG datasets, demonstrating the approach's robustness across different clinical contexts while maintaining the benefits of supervised contrastive learning.

The supervised contrastive loss \( L_{\text{con}} \) enhances feature discrimination by drawing together embeddings of similar ECG signals (positive pairs) while pushing apart embeddings of dissimilar signals (negative pairs). Unlike the self-supervised approach in SimCLR~\cite{chen2020simple}, which generates positive pairs through data augmentation, our supervised contrastive loss explicitly utilizes class label information. It contrasts embeddings from the same class (positive set) against embeddings from different classes in the batch (negative set). The supervised contrastive loss is formulated as:

\begin{equation}
L_{\text{con}} = -\sum_{i \in I} \frac{1}{|P(i)|} \sum_{p \in P(i)} \log \frac{\exp\left(\text{sim}(Z_i, Z_p) / \tau\right)}{\sum_{j \in A(i)} \exp\left(\text{sim}(Z_i, Z_j) / \tau\right)}, \tag{5}
\end{equation}
where $Z_i$ denotes the encoder output (embedding) for the $i$-th sample, and $Z_p$ represents the embeddings of samples that share the same class label as sample $i$. The set $A(i)$ is defined as all batch indices excluding $i$, that is, $A(i) = I \setminus \{i\}$. The set of positive indices $P(i)$ includes all indices $p \in A(i)$ such that $y_p = y_i$, i.e., $P(i) = \{p \in A(i) : y_p = y_i\}$. The parameter $\tau$ is a temperature scalar that controls the sensitivity of similarity measurements. For measuring relationships between embedding vectors, we employ the cosine similarity metric from among several possible similarity functions. The cosine similarity function $\text{sim}(Z_1, Z_2)$ that quantifies embedding alignment is defined as:
\begin{equation}
\text{sim}(Z_1, Z_2) = \frac{Z_1^\top Z_2}{\|Z_1\| \|Z_2\|},\tag{6}
\end{equation}
where $Z_1$ and $Z_2$ represent embedding vectors in the representation space. This similarity measure effectively captures the relationship between different embeddings, enabling the model to learn discriminative features specific to ECG classification tasks. 

The supervised contrastive loss, \( L_{\text{con}} \), ensures that embeddings of ECG signals sharing the same class label are pulled closer together. In contrast, embeddings of samples from different classes are pushed further apart. This structured optimization facilitates a well-separated representation space, improving downstream task performance, such as ECG signal classification and anomaly detection.
Simultaneously, the cross-entropy loss \( L_{{\text{cls}}} \) is used to ensure accurate class predictions. It is mathematically defined as:
\[
{L_{\text{cls}}} = - {\sum_{i \in I}}\sum_{c=1}^{C} y_c \log \hat{y}_c,\tag{7}
\]
where $y_c$ is the true label for the $i$-th sample, and $\hat{y}_c$ is the predicted probability for the sample. This loss minimizes the discrepancy between the predicted and actual labels, encouraging high-confidence predictions.
The combined ECG-NAT loss function for fine-tuning is defined as:
\begin{equation}
L_{\text{t}} = \alpha L_{\text{con}} + (1 - \alpha)  L_{{\text{cls}}},\tag{8}
\end{equation}
where \( 0 \leq \alpha \leq 1\) is a weighting parameter that controls the balance between the supervised contrastive loss and the classification loss. By optimizing \( L_{\text{t}} \), the model adapts its encoder and classifier to improve both feature extraction and classification performance on ECG-specific tasks.
This fine-tuning framework enables ECG-NAT to adapt effectively to diverse ECG classification tasks. It maintains the model's ability to extract meaningful and discriminative features from ECG signals, achieving robust performance across various downstream applications.

\section{Experimental Results} 
\label{se:Experimental Setup}
Presented in this section are comprehensive evaluations of ECG-NAT across multiple datasets, with performance comparisons against current best-performing techniques. This includes analyzing its effectiveness in low-resource scenarios and examining the impact of architectural components and hyperparameters on classification accuracy and computational efficiency.
\begin{table}[H]
\setlength{\tabcolsep}{0.2cm}
\centering
\caption{Overview of datasets for ECG-NAT model}
\label{tab:ecg_nat_datasets}
\scalebox{0.8}{
\begin{tabular}{l|lcccc}
\toprule
\textbf{Stage} & \textbf{Datasets} & \textbf{Patients} & \textbf{Records} & \textbf{Duration (s)} & \textbf{Sampling Frequency (Hz)} \\
\midrule
Pretraining & Chapman \cite{ChapmanECG2020} & 10,646 & 10,646 & 10s & 500 Hz \\[2pt]
             & Ningbo \cite{zheng2020optimal} & 45,152 & 34,905 & 10s & 500 Hz \\[2pt]
\midrule
Fine-tuning  & PTB-XL \cite{wagner2020ptb} & 18,885 & 21,837 & 10s & 500 Hz \\[2pt]
             & CPSC2018 \cite{liu2018open} & 9,831 & 6,877 & 6-60s & 500 Hz \\
\bottomrule
\end{tabular}
}
\end{table}
\subsection{Datasets and Evaluation metrics}
The ECG-NAT framework was developed and validated using several standard ECG datasets, which are summarized in Table~\ref{tab:ecg_nat_datasets}. Our approach comprises two main phases: self-supervised pre-training, followed by supervised fine-tuning, each utilizing specific datasets and loss functions. For the initial self-supervised pretraining, two extensive unlabeled datasets, Chapman \cite{ChapmanECG2020} and Ningbo \cite{zheng2020optimal}, were used to learn robust and generalizable ECG representations from diverse signal patterns. Subsequently, for supervised fine-tuning and performance evaluation on specific classification tasks, the widely recognized benchmark datasets PTB-XL \cite{wagner2020ptb} and CPSC2018 \cite{liu2018open} were employed.  
For each dataset, we randomly split the samples into 80\% for training and 20\% for testing. To account for variability due to random splits, this process was repeated five times, and the reported performance metrics are the averages across these runs. This division makes sure that the model's ability to generalize to new data is thoroughly tested.  
For the fine-tuning phase, we adopt a multi-class classification framework on PTB-XL and CPSC2018. Both datasets contain recordings with multiple concurrent diagnostic labels, which presents a challenge for multi-class classification. Following standard preprocessing protocols \cite{na2024guiding,zhang2022maefe}, we retain only single-label instances to ensure a well-defined multi-class setting. This filtering approach maintains consistency with prior benchmarking studies while enabling direct comparison with existing methods on the same datasets.
Table~\ref{tab:ecg_nat_datasets} offers a comprehensive summary of all datasets employed in our methodology, while Table~\ref{tab:ptb_table} and Table~\ref{tab:cpsc_table} present detailed diagnostic distributions for PTB-XL and CPSC2018, respectively.
\begin{table}[H]
\setlength{\tabcolsep}{0.5cm}
\centering
\caption{Distribution of ECG Record Counts in the PTB-XL Dataset}
\label{tab:ptb_table}
\scalebox{0.9}{
\begin{tabular}{@{}llc@{}}
\toprule
\textbf{Superclass} & \textbf{Description} & \textbf{Original} \\ 
\midrule
NORM & Normal ECG & 9528 \\ 
MI & Myocardial Infarction & 5486 \\ 
STTC & ST/T Change & 5250 \\ 
CD & Conduction Disturbance & 4907 \\ 
HYP & Hypertrophy & 2655 \\ 
\bottomrule
\end{tabular}
}
\end{table}
\begin{table}[H]
\setlength{\tabcolsep}{0.5cm}
\centering
\caption{Summary of ECG Conditions and Instance Counts in the CPSC2018 Dataset}
\label{tab:cpsc_table}
\scalebox{0.9}{
\begin{tabular}{@{}llc@{}}
\toprule
\textbf{Type} & \textbf{Description} & \textbf{Original} \\ 
\midrule
N & Normal & 918 \\
AF & Atrial fibrillation & 1098 \\
I-AVB & 1st-degree atrioventricular block & 704 \\
LBBB & Left bundle branch block & 207 \\
RBBB & Right bundle branch block & 1695 \\
PAC & Premature atrial contraction & 574 \\
PVC & Premature ventricular contraction & 653 \\
STD & ST-segment depression & 826 \\
STE & ST-segment elevated & 202 \\
\bottomrule
\end{tabular}
}
\end{table}
The pre-training phase of our evaluation utilizes Mean Squared Error (MSE). A thorough assessment of the fine-tuned model's performance on ECG-NAT considers Accuracy (Acc) for direct predictive correctness. To guarantee an equitable performance evaluation, especially with imbalanced data, F1-score (F1) and Area Under the Curve (AUC) are also included. The terms TP, TN, FP, and FN refer to the corresponding numbers of true positives, true negatives, false positives, and false negatives. The fine-tuning process relies on a dual-loss framework that merges cross-entropy classification with supervised contrastive loss, aiming to extract distinguishing features and enhance class prediction precision. Evaluation metrics are detailed below:
\begin{flalign}
Accuracy &= \frac{TP+TN}{TP+TN+FP+FN}, \tag{9}
\end{flalign}

\begin{flalign}
F1 &= \frac{1}{class} \sum_{i} 2 \times \frac{precision_i \times recall_i}{precision_i + recall_i}, \tag{10}
\end{flalign}

\begin{flalign}
AUC &= \int_{0}^{1} TPR(FPR^{-1}(t))dt, \tag{11}
\end{flalign}
where:
\begin{flalign}
precision &= \frac{TP}{TP+FP} \quad , \quad  recall = \frac{TP}{TP+FN},\nonumber
\end{flalign}
\begin{flalign}
TPR &= \frac{TP}{TP+FN} \quad , \quad FPR = \frac{FP}{FP+TN}.\nonumber
\end{flalign}
This approach enables a comprehensive evaluation of the model's performance during fine-tuning across various datasets.
\subsection{Setting}
The ECG-NAT model architecture employs a self-supervised neighborhood attention transformer for ECG representation learning, as detailed in Table~\ref{tab:model_params}.  The pre-processing and normalization of 12-lead ECG signals, comprising 2500 time points per lead, initiates our pipeline, followed by their passage through several precisely crafted steps.  During the tokenization phase, pre-processed signals are transformed into significant token embeddings by two Conv1d layers. The encoding stage then employs four specialized NAT blocks, each with a precise configuration: an embedding dimension of 96, variable attention heads, and an MLP ratio of 4.0. Essential for hierarchical representation are the down-sampler modules, composed of three Conv1d layers, each with ReLU activation, as they reduce spatial dimensions and increase feature channels after the NAT blocks. A fully connected layer handles the final classification tasks, while the decoder's ConvTranspose1d layers reconstruct the original signal during pretraining. The complete ECG-NAT model utilizes the NAT-Tiny configuration, which contains approximately 30 million parameters.

In the training phase, Gaussian noise is introduced into the tokenized input embeddings. This is performed by introducing minor alterations that follow a normal distribution with a standard deviation of 0.2. This noise injection helps the model acquire more general and robust representations while preserving the fundamental structure of the original ECG signals. Moreover, the temperature parameter $\tau$ is utilized within the supervised contrastive loss to modulate similarity scores, thereby refining the sharpness of the loss landscape.
 Our system was developed using PyTorch and our custom NATTEN package, which provides optimized C++ and CUDA kernels for enhanced computational speed.  The model training employed cosine annealing learning rate scheduling in conjunction with the AdamW optimizer.  All trials were conducted on an Ubuntu 22.04 LTS system, utilizing a single RTX 2080Ti GPU with 12 GB of memory.

\begin{table*}[]
\setlength{\tabcolsep}{0.4cm}
\centering
\caption{ECG-NAT model hyperparameter configuration information}
\scalebox{0.7}{
\begin{tabular}{@{}llll@{}}
\toprule
\textbf{Stage} & \textbf{Module} & \textbf{Layer Details} & \textbf{Feature Shape} \\
\midrule
 & Time Domain Input & - & 12 $\times$ 2500 \\
\midrule
Tokenizer & Conv1d $\times$ 2 & $K_{\text{s}}$ = 3,3, $st$ = 2,2, $P$ = 0,0 & 48 $\times$ 1250 \\
 &  &  & 92 $\times$ 625\\
\midrule
\addlinespace[4pt]
\multirow{3}{*}{Encoder} & NAT Block $\times$ 4 & emb = 96, n-head = (2,4,8,16), mlp-ratio = 4.0 & 92 $\times$ 625 \\
 &  &  & 192 $\times$ 312 \\
 & Conv1d $\times$ 3 & $K_{\text{s}}$ = 2,2,2, $st$ = 2,2,2, $P$ = 0,0,0 & 384 $\times$ 156 \\
 &  &  & 768 $\times$ 78 \\
\midrule
\multirow{4}{*}{Decoder} & ConvTranspose1d $\times$ 1 & $K_{\text{s}}$ = 3, $st$ = 2, $P$ = 0 & 768 $\times$ 78 \\
 &  &  & 384 $\times$ 156 \\
 
 & ConvTranspose1d $\times$ 4 & $K_{\text{s}}$ = 2,2,2,4, $st$ = 2,2,2,4, $P$ = 0,0,0,0 & 192 $\times$ 312 \\
 &  &  & 92 $\times$ 625 \\
 &  &  & 48 $\times$ 1250 \\
\midrule
Classifier & Fully-Connected & Linear(768 $\times$ 78, n-classes) & n-classes \\ 
\bottomrule
\end{tabular}
}
\label{tab:model_params}
\end{table*}
\subsection{Method Comparisons}
In this work, ECG-NAT is compared against state-of-the-art methods to evaluate its performance. The comparison frameworks include established techniques in the field and self-supervised learning (SSL) methods.
\begin{itemize}

\item  \textbf{MOCO V3}~\cite{chen2021empirical} employs a self-supervised vision transformer with contrastive learning, proving that such models match or exceed traditional supervised approaches in complex feature extraction tasks.

\item  \textbf{CMSC}~\cite{oh2022lead} enhances model reliability and value by mitigating noise-related estimation loss. It achieves this through learning from distinctions in cardiac signals across spatial, temporal, and patient dimensions.

\item  \textbf{MaeFE}~\cite{zhang2022maefe}, a family of masked autoencoders based on the Vision Transformer (ViT) architecture, is designed for self-supervised learning of ECG signals using three masking strategies: temporal (MTAE), spatial (MLAE), and combined (MLTAE).

\item  \textbf{CRT}~\cite{zhang2023self} is a self-supervised approach for learning time series representations. It operates by converting signals into temporal and spectral domains and subsequently dropping patches from both. An associated Transformer encoder then reconstructs the initial data from these dropped inputs, utilizing inter-domain correlations and an instance-level discrimination criterion to produce robust and distinct representations.

\item  \textbf{ASTCL}~\cite{wang2023adversarial} (The Adversarial Spatiotemporal Contrastive Learning) functions as a self-supervised learning paradigm for ECG signals that employs noise-based augmentations and adversarial training. It learns robust spatiotemporal representations through patient-level positive pairs without negative examples. This approach significantly improves classification performance even with minimal labeled data.

\item  \textbf{ST-MEM}~\cite{na2024guiding} captures spatial and temporal relationships in electrocardiograms through masking and lead-specific processing, excelling in arrhythmia classification after fine-tuning various ECG datasets.

\item  \textbf{DLA}~\cite{liu2024direct}, a self-supervised learning technique, pre-trains models using unlabeled multi-lead ECG data. To improve lead-specific features, it uses an asymmetric architecture that combines both single-lead and multi-lead encoders. It does this by performing contrastive learning between the representations from the two types of encoders. Subsequently, a classifier is incorporated for follow-up tasks and optimized to classify ECGs for the diagnosis of cardiovascular conditions accurately. Even with restricted labeled data, this approach yields scalable and highly effective models, rendering them well-suited for diverse Internet of Medical Things (IoMT) applications.

\item \textbf{LFBT}~\cite{liu2025lead} (Lead-Fusion Barlow Twins) is a self-supervised method for multi-lead ECGs that fuses intra- and inter-lead Barlow Twins losses. It avoids negative pairs and large batches, enabling efficient fusion of all leads during pretraining.

\item \textbf{JAMC}~\cite{ge2025jamc} (Jigsaw-based Autoencoder with Masked Contrastive Learning) unifies contrastive and jigsaw-based reconstruction learning. It applies lead permutation and masking to form view pairs, learning both global invariances and local morphological details from 12-lead ECGs.
\end{itemize}

\subsection{Method Comparison Results}
The experimental evaluation assesses ECG-NAT against other state-of-the-art self-supervised learning methods using two cases of linear evaluation and fine-tuning approaches. The results presented here are sourced directly from the original papers of each comparative method under similar experimental circumstances, including consistent pre-processing protocols, resampling to 2500 Hz, and comparable self-supervised architectural frameworks to ensure rigorous and standardized benchmarking.
In linear evaluation, all pre-trained model parameters remain frozen except for the final classification layer, which directly assesses the quality of learned representations during masked autoencoder pretraining by effectively employing ECG-NAT as a feature extractor. In contrast, fine-tuning optimizes the entire model architecture for downstream tasks by unfreezing the encoder that was initially pre-trained with the masked autoencoder technique. This approach allows backpropagation through the entire model with contrastive loss while simultaneously training the linear classifier, enabling the model to adapt its representations from the masked autoencoder pretraining specifically for the classification task.

As shown in Table~\ref{table:Comparisons}, under the linear evaluation protocol, where all pre-trained model parameters remain frozen except for the final classification layer, ECG-NAT demonstrates superior performance compared to other self-supervised methods. On the PTB-XL dataset, it achieves an accuracy of $0.800 \pm 0.032$, an F1-score of $0.799 \pm 0.003$, and an AUROC of $0.960 \pm 0.001$. On CPSC2018, ECG-NAT attains an accuracy of $0.782 \pm 0.004$, an F1-score of $0.855 \pm 0.005$, and an AUROC of $0.965 \pm 0.005$, outperforming all baseline methods across both datasets. These results highlight the quality of representations learned during masked autoencoder pretraining as a robust foundation for downstream tasks.

When fine-tuned and evaluated on PTB-XL and CPSC2018 (see Table~\ref{table:Comparisons2}), ECG-NAT further improves upon these results, surpassing other self-supervised approaches such as CRT~\cite{zhang2023self} and ST-MEM~\cite{na2024guiding}. Specifically, on PTB-XL, it achieves an accuracy of $0.902 \pm 0.002$, an F1-score of $0.904 \pm 0.009$, and an AUROC of $0.977 \pm 0.002$. On CPSC2018, it attains an accuracy of $0.903 \pm 0.005$, an F1-score of $0.895 \pm 0.007$, and an AUROC of $0.986 \pm 0.001$. These consistent improvements demonstrate that ECG-NAT not only extracts high-quality representations during pretraining but also effectively adapts them to specific classification tasks, outperforming contemporary methods. 
\begin{table*}[]
\setlength{\tabcolsep}{0.35cm}
\centering
\caption{\justifying Performance comparison of self-supervised learning methods for ECG classification using linear evaluation, measured by Accuracy, Macro F1-score, and AUROC}
\label{table:Comparisons}
\scalebox{0.6}{
\begin{tabular}{l|ccc|ccc}
\toprule
\textbf{ Method} & \multicolumn{3}{c}{\textbf{PTB-XL}} & \multicolumn{3}{c}{\textbf{CPSC2018}} \\
\midrule
 \textbf{} & \textbf{Accuracy} & \textbf{F1-score} & \textbf{AUROC} & \textbf{Accuracy} & \textbf{F1-score} & \textbf{AUROC} \\
\midrule
MOCOV3(2021) \cite{chen2021empirical} & 0.552 ± 0.000 & 0.142 ± 0.000 & 0.739 ± 0.006 & 0.268 ± 0.055 & 0.080 ± 0.038 & 0.712 ± 0.054 \\[2pt]
CMSC(2021) \cite{kiyasseh2021clocs} & 0.681 ± 0.032 & 0.441 ± 0.058 & 0.797 ± 0.038 & 0.361 ± 0.005 & 0.238 ± 0.022 & 0.724 ± 0.013 \\[2pt]
MTAE(2022) \cite{zhang2022maefe} & 0.683 ± 0.008 & 0.437 ± 0.012 & 0.807 ± 0.006 & 0.486 ± 0.012 & 0.349 ± 0.034 & 0.818 ± 0.010 \\[2pt]
MLAE(2022) \cite{zhang2022maefe} & 0.649 ± 0.008 & 0.382 ± 0.020 & 0.779 ± 0.008 & 0.443 ± 0.014 & 0.263 ± 0.021 & 0.794 ± 0.016 \\[2pt]
ASTCL(2024) \cite{wang2023adversarial} & - & - & 0.823 ± 0.006 & - & - & 0.842 ± 0.046 \\[2pt]
ST-MEM(2024) \cite{na2024guiding} & \underline{0.726 ± 0.005} & \underline{0.508 ± 0.008} & \underline{0.838 ± 0.011} & \underline{0.723 ± 0.008} & \underline{0.641 ± 0.010} & \underline{0.938 ± 0.002} \\[2pt]

\midrule
ECG-NAT (ours) & \textbf{0.800 ± 0.032} & \textbf{0.799 ± 0.003} & \textbf{0.960 ± 0.001} & \textbf{0.782 ± 0.004} & \textbf{0.855 ± 0.005} & \textbf{0.965 ± 0.005} \\[2pt]
\bottomrule
\end{tabular}
}
\end{table*}
\begin{table*}[]
\setlength{\tabcolsep}{0.35cm}
\centering
\caption{\justifying Performance comparison of self-supervised learning methods for ECG classification after fine-tuning, evaluated using Accuracy, Macro F1-score, and AUROC}
\label{table:Comparisons2}
\scalebox{0.6}{
\begin{tabular}{l|ccc|ccc}
\toprule
\textbf{Method} & \multicolumn{3}{c}{\textbf{PTB-XL}} & \multicolumn{3}{c}{\textbf{CPSC2018}} \\
\midrule
 \textbf{} & \textbf{Accuracy} & \textbf{F1-score} & \textbf{AUROC} & \textbf{Accuracy} & \textbf{F1-score} & \textbf{AUROC} \\
\midrule
MOCO V3(2021) \cite{chen2021empirical} & 0.799 ± 0.004 & 0.644 ± 0.010 & 0.913 ± 0.002 & 0.852 ± 0.002 & 0.838 ± 0.002 & 0.967 ± 0.003 \\[2pt]
CMSC(2021) \cite{kiyasseh2021clocs} & 0.724 ± 0.067 & 0.510 ± 0.115 & 0.877 ± 0.003 & 0.736 ± 0.006 & 0.717 ± 0.006 & 0.938 ± 0.006 \\[2pt]
MTAE(2022) \cite{zhang2022maefe} & 0.789 ± 0.002 & 0.613 ± 0.015 & 0.910 ± 0.001 & 0.793 ± 0.004 & 0.769 ± 0.004 & 0.961 ± 0.001 \\[2pt]
MLAE(2022) \cite{zhang2022maefe} & 0.802 ± 0.004 & 0.625 ± 0.009 & 0.915 ± 0.001 & 0.834 ± 0.007 & 0.816 ± 0.009 & 0.973 ± 0.002 \\[2pt]
CRT(2023) \cite{zhang2023self} & \underline{0.878 ± 0.029} & 0.684 ± 0.058 & 0.892 ± 0.007 & - & - & - \\[2pt]
ASTCL(2024) \cite{wang2023adversarial} & - & 0.572 ± 0.077 & - & - & 0.631 ± 0.067 & - \\[2pt]
ST-MEM(2024) \cite{na2024guiding} & 0.825 ± 0.002 & \underline{0.655 ± 0.003} & \underline{0.933 ± 0.003} & \underline{0.872 ± 0.009} & \underline{0.857 ± 0.012} & \underline{0.980 ± 0.001} \\[2pt]
DLA(2025) \cite{liu2024direct} & 0.779 ± 0.007 & 0.624 ± 0.003 & 0.901 ± 0.000 & 0.706 ± 0.002 & 0.670 ± 0.001 & 0.931 ± 0.000 \\[2pt]

LEBT(2025) \cite{liu2025lead} & -- & -- & 0.908 ± 0.000 & -- & -- & 0.936 ± 0.000 \\[2pt]

JAMC(2025) \cite{ge2025jamc} & -- & -- & -- & 0.706 ± 0.000 & 0.666 ± 0.000 & 0.915 ± 0.000 \\[2pt]
\midrule
ECG-NAT (ours) & \textbf{0.902 ± 0.002} & \textbf{0.904 ± 0.009} & \textbf{0.977 ± 0.002} & \textbf{0.903 ± 0.005} & \textbf{0.895 ± 0.007} & \textbf{0.986 ± 0.001} \\[2pt]
\bottomrule
\end{tabular}
}
\end{table*}
\subsection{Low-Resource Classification Performance Results}
ECG-NAT provides exceptional benefits in low-resource clinical environments where labeled ECG data is scarce. This scarcity is a common challenge in healthcare settings due to the multitude of heart diseases and the limited availability of skilled cardiologists. As shown in Table~\ref{table:LowResource}, ECG-NAT was evaluated across different amounts of training data (1\%, 5\%, and 100\%) from each dataset. The model maintained outstanding stability under these extremely constrained circumstances, even as conventional supervised techniques exhibited notable performance drops.

Using only 1\% of the training data on the PTB-XL dataset, ECG-NAT achieved an AUROC of $0.880 \pm 0.002$. This demonstrates that it is very competitive with, or even better than, most other methods that use 5\% of the data. For example, ST-MEM (2024) achieved an AUROC of $0.878 \pm 0.011$ with $5\%$ of the data, but ECG-NAT performed just as well or better with significantly less data.
ECG-NAT reaches an AUROC of $0.931 \pm 0.001$ with only $1\%$ of the training data. This is much better than any other method tested in the same low-resource conditions. The $0.952 \pm 0.004$ AUROC achieved by ST-MEM(2024) with five times as much training data (5\%) is very close to this result.
These results demonstrate that ECG-NAT consistently outperforms other self-supervised methods, using different amounts of training data, and shows superior capability in identifying important ECG patterns even under severely limited labeled data conditions.
\begin{table*}[]
	\setlength{\tabcolsep}{0.35cm}
	\centering
	\caption{\justifying Experimental results under low-resource conditions (1\% and 5\% training data) compared to full training data (100\%), using consistent test sets and measuring AUROC scores across 12-lead ECG signals (averaged over three random samplings).}
	\label{table:LowResource}
	\scalebox{0.6}{
		\begin{tabular}{l|ccc|ccc}
			\toprule
			\textbf{Method} & \multicolumn{3}{c}{\textbf{PTB-XL}} & \multicolumn{3}{c}{\textbf{CPSC2018}} \\
			\midrule
			& \textbf{1\%} & \textbf{5\%} & \textbf{100\%} & \textbf{1\%} & \textbf{5\%} & \textbf{100\%} \\
			\midrule
			MOCO V3(2021) \cite{chen2021empirical} & $0.797 \pm 0.006$ & $0.826 \pm 0.015$ & $0.913 \pm 0.002$ & $0.791 \pm 0.045$ & $0.903 \pm 0.019$ & $0.967 \pm 0.003$ \\[2pt]
			CMSC(2021) \cite{kiyasseh2021clocs} & $0.648 \pm 0.064$ & $0.773 \pm 0.023$ & $0.877 \pm 0.003$ & $0.625 \pm 0.013$ & $0.732 \pm 0.038$ & $0.938 \pm 0.006$ \\[2pt]
			MTAE(2022) \cite{zhang2022maefe} & $0.707 \pm 0.024$ & $0.713 \pm 0.001$ & $0.910 \pm 0.001$ & $0.670 \pm 0.032$ & $0.756 \pm 0.013$ & $0.961 \pm 0.001$ \\[2pt]
			MLAE(2022) \cite{zhang2022maefe} & $0.793 \pm 0.007$ & $0.838 \pm 0.018$ & $0.915 \pm 0.001$ & $0.860 \pm 0.013$ & $0.922 \pm 0.007$ & $0.973 \pm 0.002$ \\[2pt]
			ST-MEM(2024) \cite{na2024guiding} & $\underline{0.815 \pm 0.012}$ & $\underline{0.878 \pm 0.011}$ & $\underline{0.933 \pm 0.003}$ & $\underline{0.897 \pm 0.025}$ & $\underline{0.952 \pm 0.004}$ & $\underline{0.980 \pm 0.001}$ \\[2pt]
            DLA(2025) \cite{liu2024direct} & 0.789 ± 0.002 & - & 0.920 ± 0.000 & 0.823 ± 0.002 & - & 0.945 ± 0.003 \\[2pt]
                \midrule
			ECG-NAT (ours) & $\mathbf{0.880 \pm 0.001}$ & $\mathbf{0.927 \pm 0.005}$ & $\mathbf{0.977 \pm 0.002}$ & $\mathbf{0.931 \pm 0.001}$ & $\mathbf{0.961 \pm 0.002}$ & $\mathbf{0.986 \pm 0.001}$ \\[2pt]
			\bottomrule
		\end{tabular}
	}
\end{table*}
\subsection{Ablation Study}
This section describes the effects of major design changes on ECG-NAT's performance, as evaluated through ablation experiments focusing on four key components: Gaussian noise masking, the contrastive loss function, and NAT transformer architecture compared to alternative transformer variants. These components were separately tested on the PTB-XL and CPSC2018 datasets, with results detailed in Table~\ref{tab:ablation_study}.
\begin{table}[H]
	\setlength{\tabcolsep}{0.4cm}
	\centering
	\caption{\justifying Ablation study on two components, contrastive loss and masking, assessed in terms of Accuracy, F1-score, and AUROC}
	\label{tab:ablation_study}
	\scalebox{0.7}{
		\begin{tabular}{l|l|ccc}
			\toprule
			\textbf{Dataset} & \textbf{Method} & \textbf{Accuracy} & \textbf{F1-score} & \textbf{AUROC} \\
			\midrule
			\multirow{6}{*}{PTB-XL} & Zero-mask & 0.855 $\pm$ 0.002 & 0.856 $\pm$ 0.002 & 0.955 $\pm$ 0.001 \\[2pt]
			& No-Contrastive Loss & 0.875 $\pm$ 0.002 & 0.876 $\pm$ 0.002 & 0.965 $\pm$ 0.001 \\[2pt]
			& ECG-ViT & 0.775 $\pm$ 0.000 & 0.758 $\pm$ 0.000 & 0.906 $\pm$ 0.000 \\[2pt]
			& ECG-Swin & 0.806 $\pm$ 0.000 & 0.791 $\pm$ 0.000 & 0.912 $\pm$ 0.000 \\[2pt]
                \cmidrule(lr){2-5}
			& ECG-NAT (Ours) & \textbf{0.902 $\pm$ 0.002} & \textbf{0.904 $\pm$ 0.009} & \textbf{0.977 $\pm$ 0.002} \\[2pt]
			\toprule
			\multirow{6}{*}{CPSC2018} & Zero-mask & 0.845 $\pm$ 0.005 & 0.839 $\pm$ 0.008 & 0.965 $\pm$ 0.001 \\[2pt]
			& No-Contrastive Loss & 0.878 $\pm$ 0.002 & 0.879 $\pm$ 0.007 & 0.975 $\pm$ 0.001 \\[2pt]
			&  ECG-ViT  & 0.794 $\pm$ 0.000 & 0.776 $\pm$ 0.000 & 0.928 $\pm$ 0.000 \\[2pt]
			& ECG-Swin & 0.834 $\pm$ 0.000 & 0.819 $\pm$ 0.000 & 0.940 $\pm$ 0.000 \\[2pt]
                \cmidrule(lr){2-5}
			& ECG-NAT (Ours) & \textbf{0.903 $\pm$ 0.005} & \textbf{0.895 $\pm$ 0.007} & \textbf{0.986 $\pm$ 0.001} \\[2pt]
			\bottomrule
		\end{tabular}
	}
\end{table}
On PTB-XL, with the incorporation of Gaussian noise masking, the ECG-NAT model performs noticeably better than the zero-mask version, suggesting improved robustness to realistic ECG signal variability. Accuracy increases by $4.7\%$ (from $0.855 \pm 0.002$ to $0.902 \pm 0.002$), and on CPSC2018, it increases by $5.8\%$ (from $0.845 \pm 0.005$ to $0.903 \pm 0.005$). The F1-score and AUROC metrics show similar improvements, supporting the conclusion that Gaussian noise masking improves representation stability under noise and perturbations commonly observed in ECG recordings.

The impact of the contrastive loss function on fine-tuning is also evident. ECG-NAT performs better than the ``No-Contrastive Loss'' version. For example, accuracy increases from $0.875 \pm 0.002$ to $0.902 \pm 0.002$ on PTB-XL and from $0.878 \pm 0.002$ to $0.903 \pm 0.005$ on CPSC2018. These results suggest that supervised contrastive learning improves the model's ability to distinguish between clinically important ECG pattern variations by enhancing class separability in the learned representation space.

\begin{figure}[H]
  \centering
  \begin{subfigure}[b]{0.48\textwidth}
    \centering
    \includegraphics[width=\textwidth]{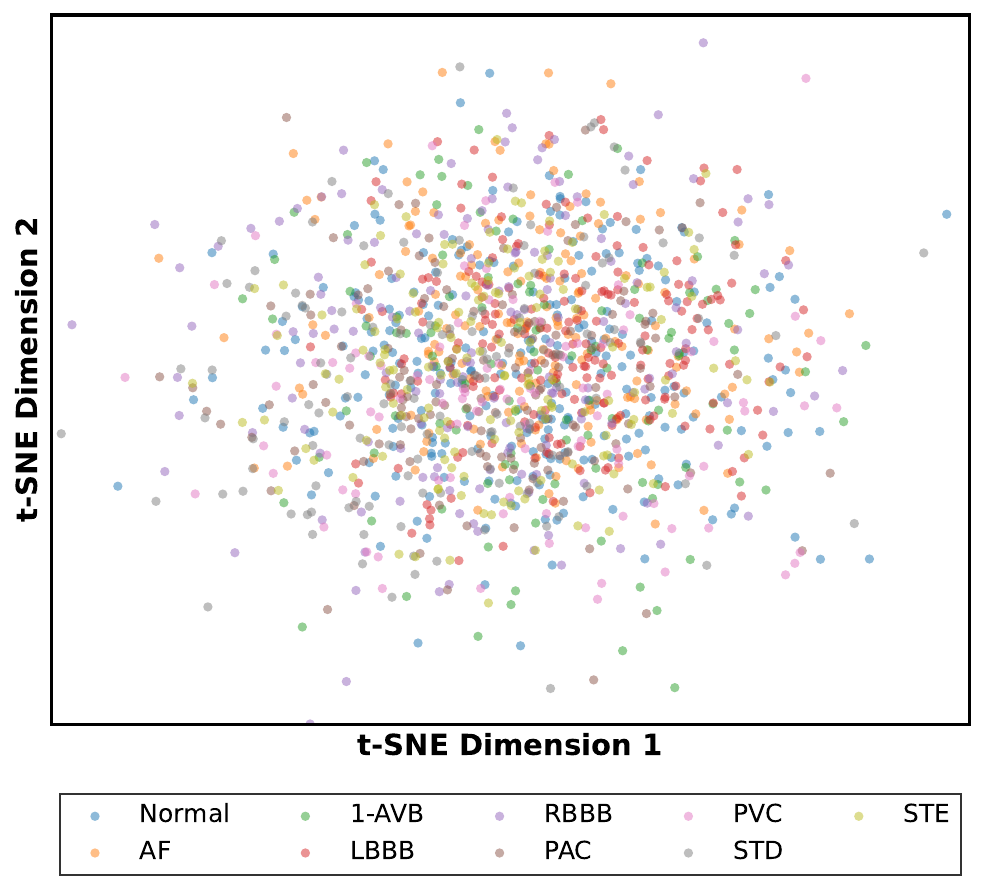}
    \caption{CPSC2018 without Contrastive Loss}
    \label{fig:cpsc_without}
  \end{subfigure}
  \hfill
  \begin{subfigure}[b]{0.48\textwidth}
    \centering
    \includegraphics[width=\textwidth]{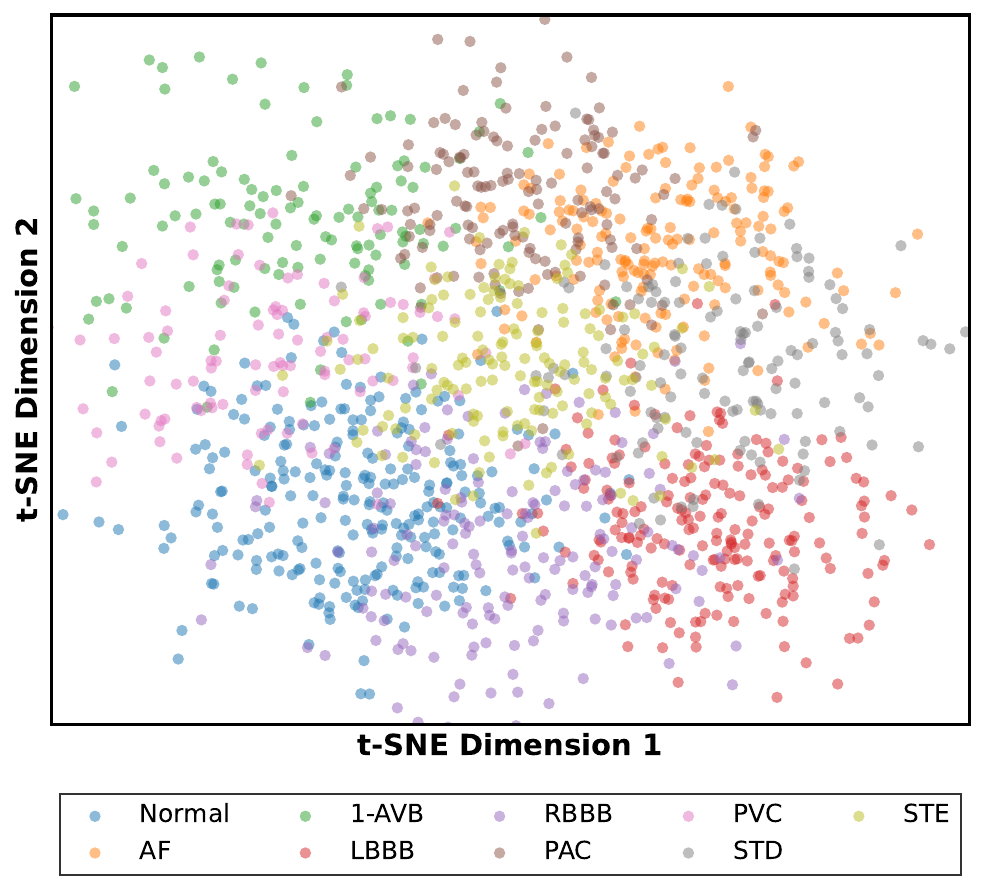}
    \caption{CPSC2018 with Supervised Contrastive Loss}
    \label{fig:cpsc_with}
  \end{subfigure}
  
  \vspace{0.5cm} 
  
  \begin{subfigure}[b]{0.48\textwidth}
    \centering
    \includegraphics[width=\textwidth]{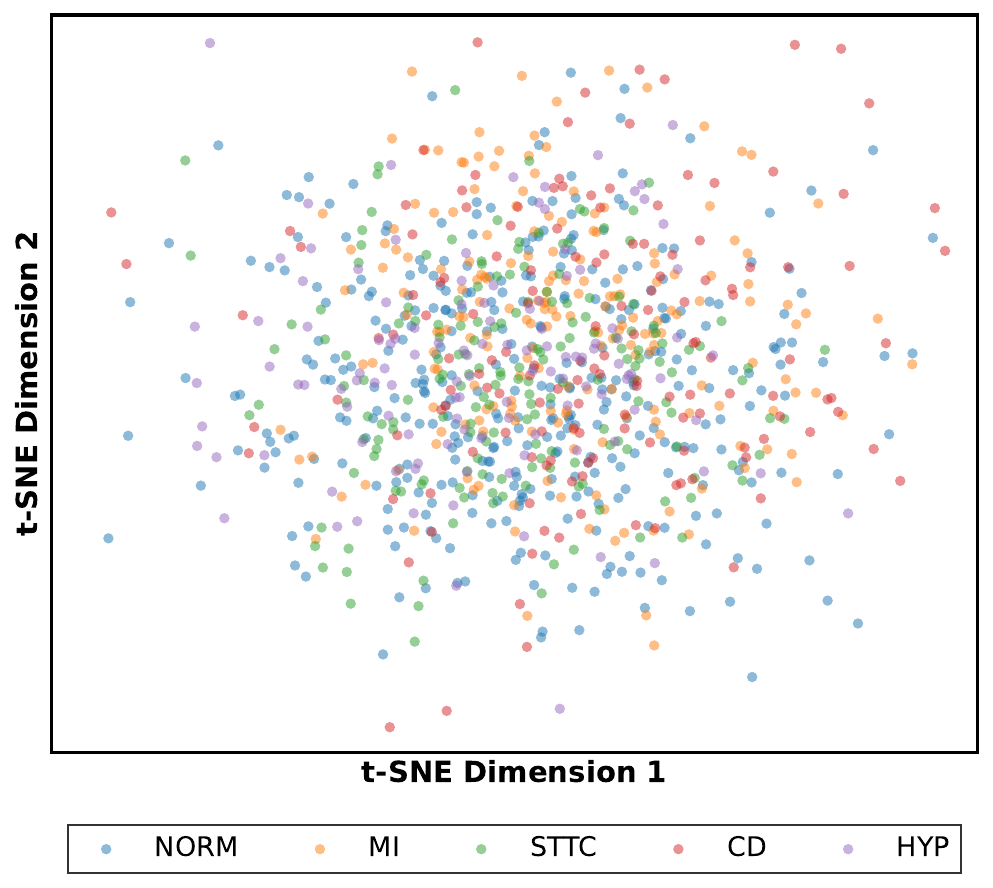}
    \caption{PTB-XL without Contrastive Loss}
    \label{fig:ptbxl_without}
  \end{subfigure}
  \hfill
  \begin{subfigure}[b]{0.48\textwidth}
    \centering
    \includegraphics[width=\textwidth]{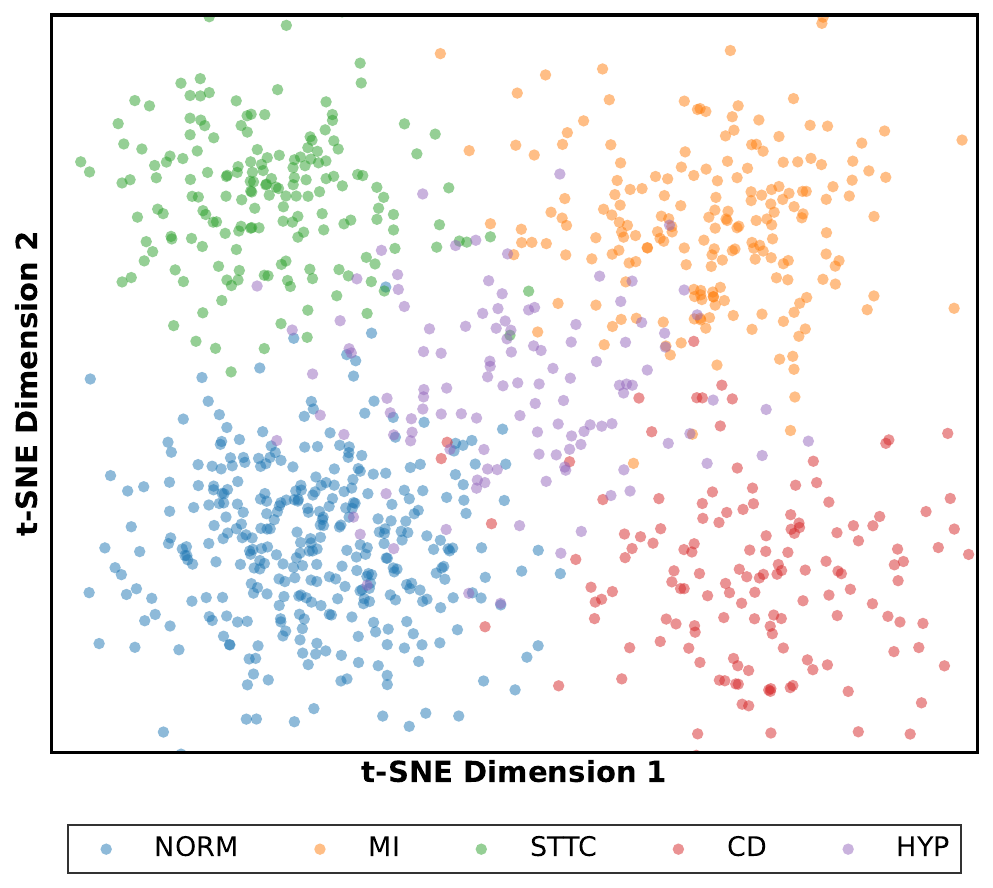}
    \caption{PTB-XL with Supervised Contrastive Loss}
    \label{fig:ptbxl_with}
  \end{subfigure}
  
\caption{\justifying t-SNE visualization of learned ECG representations with and without supervised contrastive loss. Top row (a,b): CPSC2018 dataset (9 classes) shows improved cluster separation (Silhouette: -0.185 $\rightarrow$ -0.160, Davies-Bouldin: 9.84 $\rightarrow$ 8.32). Bottom row (c,d): PTB-XL dataset (5 classes) demonstrates enhanced inter-class boundaries. Embedding quality improvements align with classification performance gains in Tables~\ref{tab:ptb_table} and~\ref{tab:cpsc_table}.}
\label{fig:tsne_visualization}
\end{figure}

{In addition, we visualize the learned embeddings using t-SNE on both CPSC2018 and 
PTB-XL test sets in Figure~\ref{fig:tsne_visualization}. Models trained with supervised contrastive loss exhibit enhanced cluster separability, achieving a 13.7\% improvement 
in Silhouette Score and a 15.4\% reduction in Davies-Bouldin Index on the CPSC2018 dataset. These embedding quality improvements align with the classification performance 
gains reported in Table~\ref{tab:ablation_study}, providing supporting evidence that contrastive learning enhances discriminative feature learning.
}

Moreover, ViT and Swin transformers were replaced in our ablation study to further validate the effectiveness of our NAT-based architecture. Specifically, we used the Tiny variants of both ViT and Swin to ensure a fair comparison with the NAT-Tiny model, maintaining parity in model size and complexity. These architectures represent commonly adopted transformer baselines in self-supervised ECG classification research. To ensure consistent evaluation, the NAT backbone was substituted with ViT and Swin transformers while preserving the same pretraining and fine-tuning pipeline. Each model underwent identical masked autoencoder pretraining followed by fine-tuning for downstream classification tasks. The results demonstrate that ECG-NAT consistently outperforms both alternatives. It improves accuracy over ViT by 12.7\% on PTB-XL (from $0.775 \pm 0.000$ to $0.902 \pm 0.002$) and 10.9\% on CPSC2018 (from $0.794 \pm 0.000$ to $0.903 \pm 0.005$). Similarly, ECG-NAT surpasses Swin Transformer on PTB-XL (9.6\%) and CPSC2018 (6.9\%), indicating that neighborhood attention is well-suited for ECG classification by emphasizing localized waveform patterns, while the hierarchical architecture enables receptive field growth to incorporate broader temporal context.

All of these components work together in our two-step training process: Gaussian noise masking enhances the robustness of feature representation during pretraining, and contrastive loss improves feature discriminability during fine-tuning. This enables ECG-NAT to transition from understanding general signal properties to identifying specific clinical differences. The full ECG-NAT model outperforms all ablated variants and baseline transformer architectures on both evaluation datasets and all metrics. This demonstrates that our architectural and methodological choices are suitable for a wide range of clinical ECG classification scenarios.

\begin{figure}[t]
    \centering
    \subcaptionbox{PTB-XL dataset parameter analysis}{
        \begin{tabular}{@{}ccc@{}}
            \includegraphics[width=0.3\textwidth]{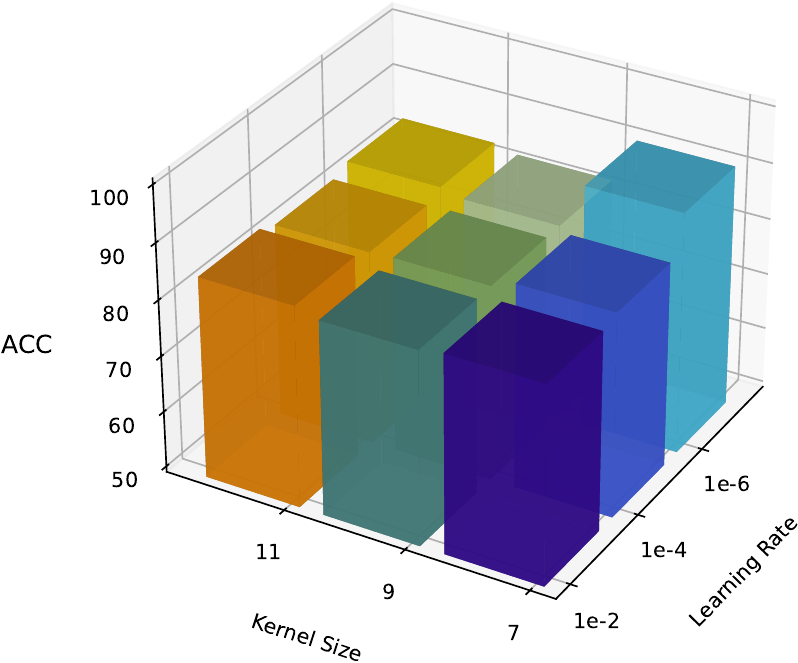} &
            \includegraphics[width=0.3\textwidth]{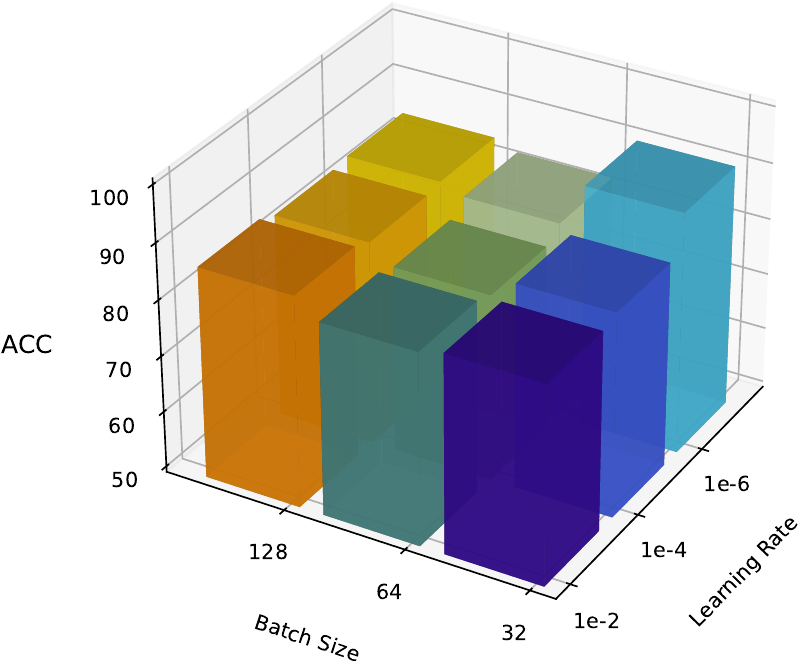} &
            \includegraphics[width=0.3\textwidth]{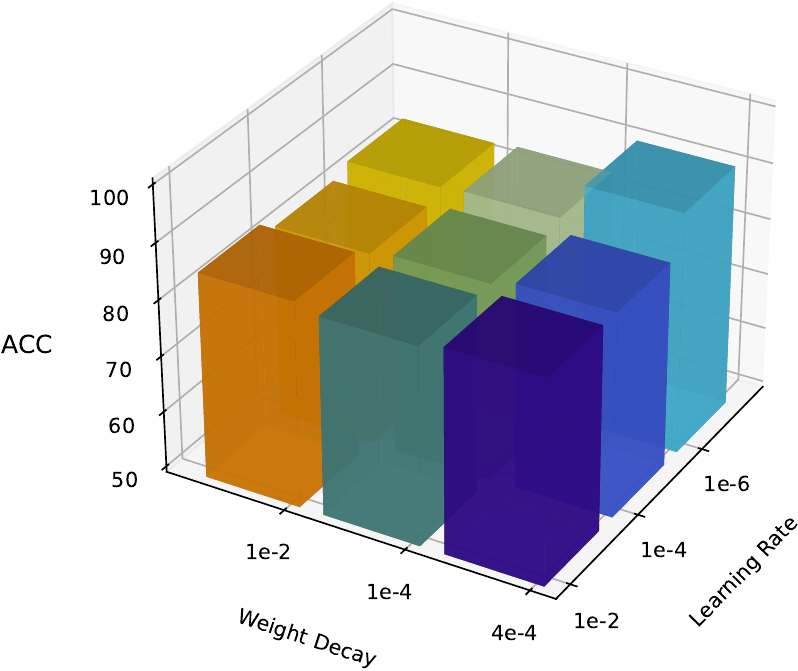}
        \end{tabular}
    }

    \vspace{1em} 

    \subcaptionbox{CPSC dataset parameter analysis}{
        \begin{tabular}{@{}ccc@{}}
            \includegraphics[width=0.3\textwidth]{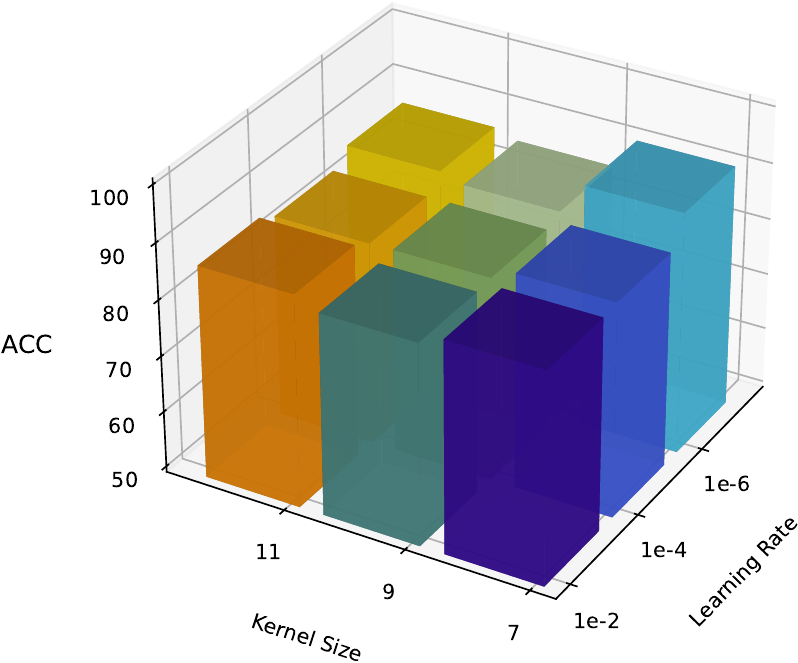} &
            \includegraphics[width=0.3\textwidth]{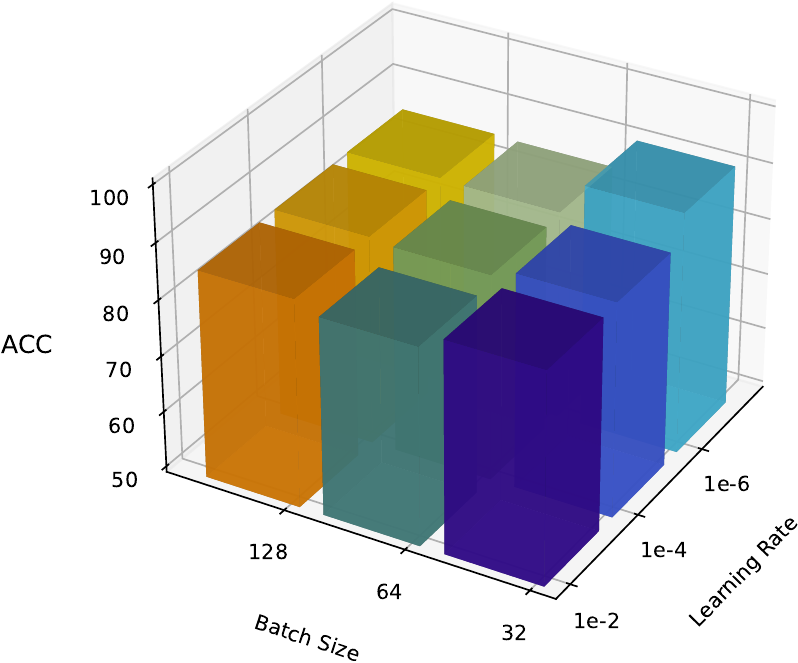} &
            \includegraphics[width=0.3\textwidth]{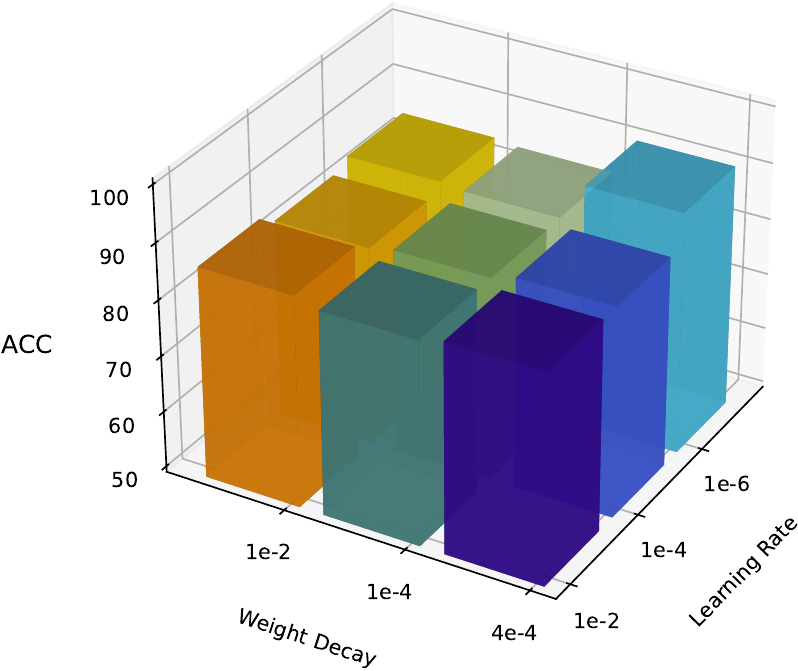}
        \end{tabular}
    }

    \caption{\justifying hyperparameter analysis showing model accuracy with different configurations of kernel size (neighborhood size), batch size, and weight decay across various learning rates for both datasets during the fine-tuning stage}
    \label{fig:hyperparameter_analysis}
\end{figure}
\subsection{Parameter Analysis}
This section examines the impact of three critical settings: kernel size (which determines the k-nearest neighbors or k-NN), batch size, and weight decay on the performance of our ECG-NAT model during the fine-tuning stage. To grasp their interconnected effects on classification accuracy, we have used 3D visualization plots. This thorough analysis has been conducted on both the PTB-XL and CPSC2018 datasets, providing a clear picture of how these hyperparameters interact to shape the model's overall effectiveness.
After exhaustively experimenting with the PTB-XL and CPSC2018 datasets (Figure~\ref{fig:hyperparameter_analysis}), we found that our best working configurations for the ECG classification model include a kernel size of 7 (corresponding to a neighborhood size), a batch size of 32, and a weight decay of $4e{-4}$. These findings are further supported by results from the CPSC2018 dataset, as shown in Figure~\ref{fig:hyperparameter_analysis}~(b), where smaller kernel sizes, equal to 7, proved better for capturing localized ECG patterns. Similarly, smaller batch sizes of 32, in contrast to larger ones such as 64 and 128, consistently outperformed across all evaluation metrics. A moderate weight decay value, such as $4e{-4}$, provides the best regularization compromise between overfitting and maintaining model capacity.

From the analysis presented in Figure~\ref{fig:hyperparameter_analysis}, the configuration with a kernel size of 7, batch size of 32, and weight decay of 4e-4 consistently achieves strong performance on both the PTB-XL and CPSC2018 datasets. The plots clearly illustrate that this setting maintains high accuracy and training stability, whereas deviations from these values lead to noticeable declines in performance. The consistent trends observed in parts(a) and (b) of Figure~\ref{fig:hyperparameter_analysis} further support the suitability of these hyperparameters for ECG classification tasks.

\begin{figure}[]
  \centering
  \begin{subfigure}[b]{0.48\textwidth}
    \raisebox{0mm}{\includegraphics[width=\textwidth]{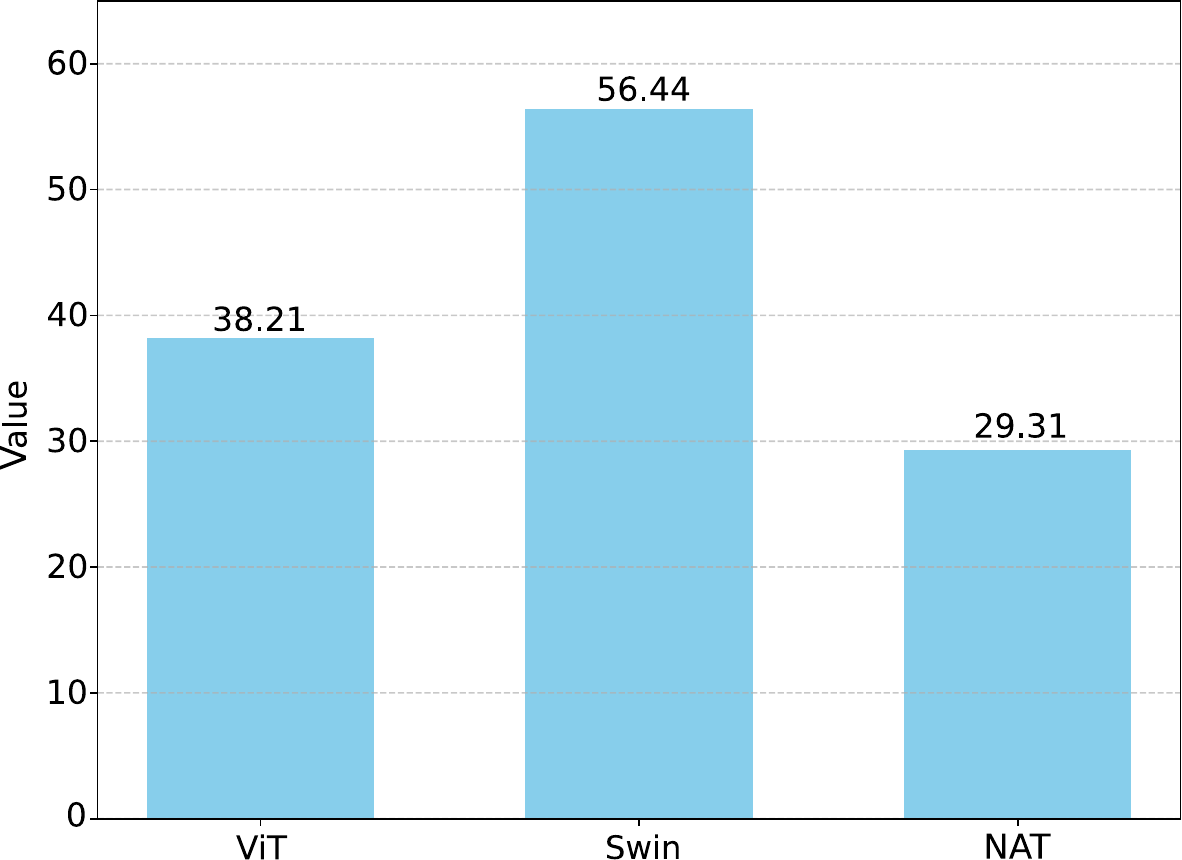}}
    \caption{Parameters (MB)}
  \end{subfigure}
  \hspace{0.02\textwidth}
  \begin{subfigure}[b]{0.48\textwidth}
    \raisebox{0mm}{\includegraphics[width=\textwidth]{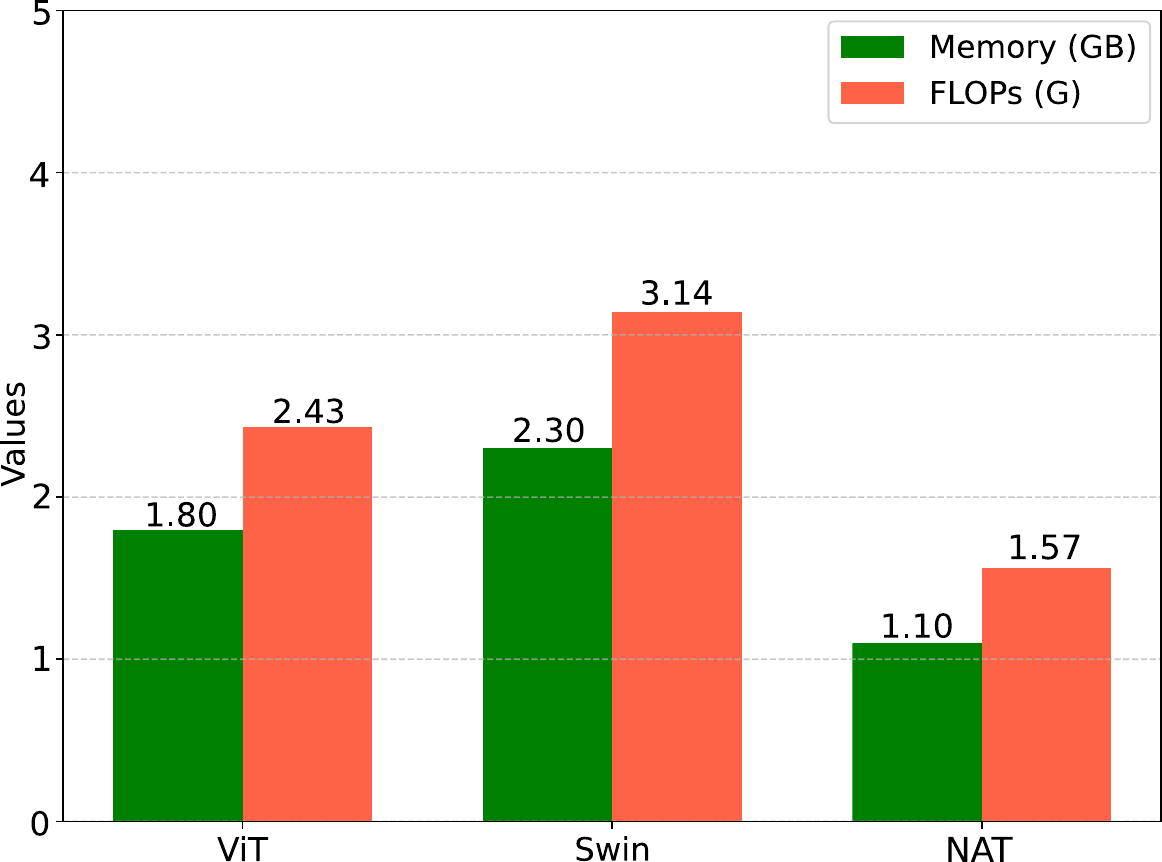}}
    \caption{Memory (GB) and FLOPs (G)}
  \end{subfigure}
  \caption{\justifying Performance efficiency analysis of ECG-NAT versus transformer baselines. (a) Model parameters in MB and (b) runtime resources, including memory (GB) and FLOPs (G), show ECG-NAT's significant computational advantages over existing architectures.}
  \label{fig:complexity}
\end{figure}
\subsection{Computational Complexity and Convergence Loss}
\begin{figure}[t]
  \centering
  \begin{subfigure}[b]{0.48\textwidth}
    \raisebox{0.1mm}{\includegraphics[width=\textwidth]{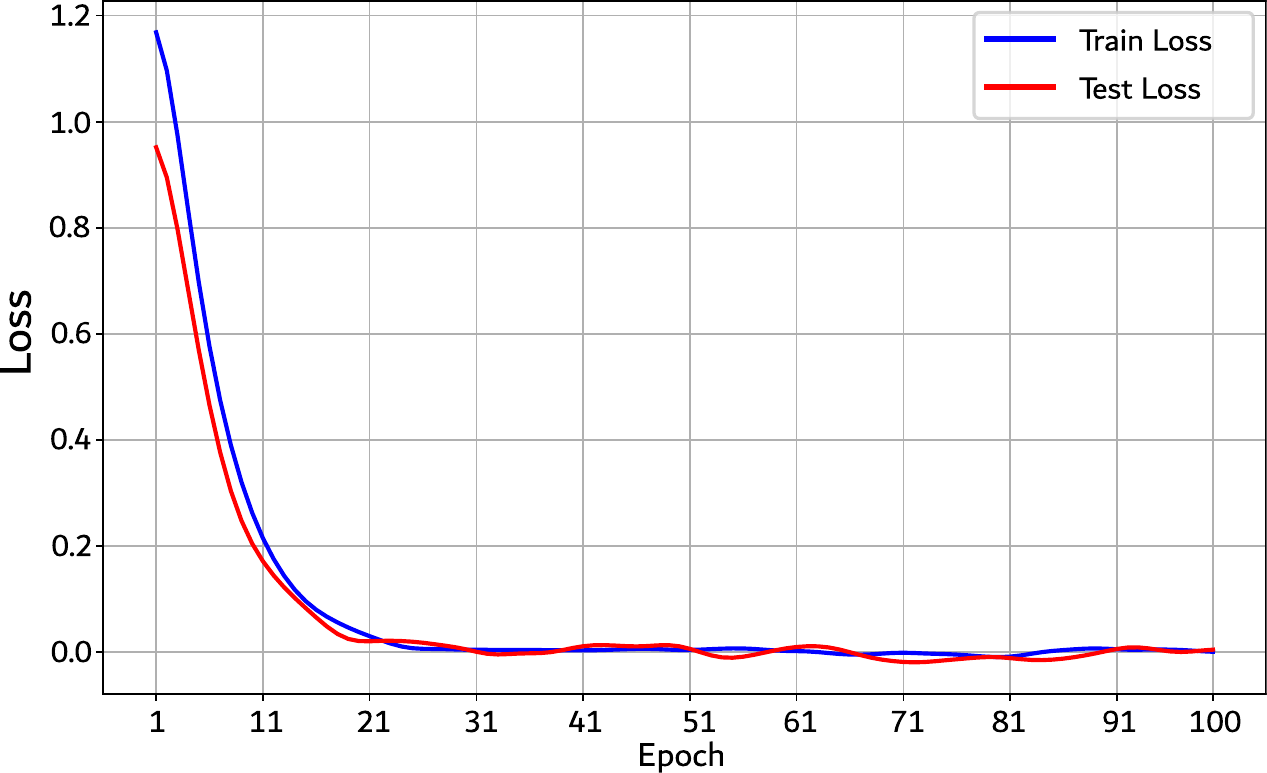}}
    \caption{Loss convergence on the PTB-XL dataset}
  \end{subfigure}
  \hfill
  \begin{subfigure}[b]{0.48\textwidth}
    \includegraphics[width=\textwidth]{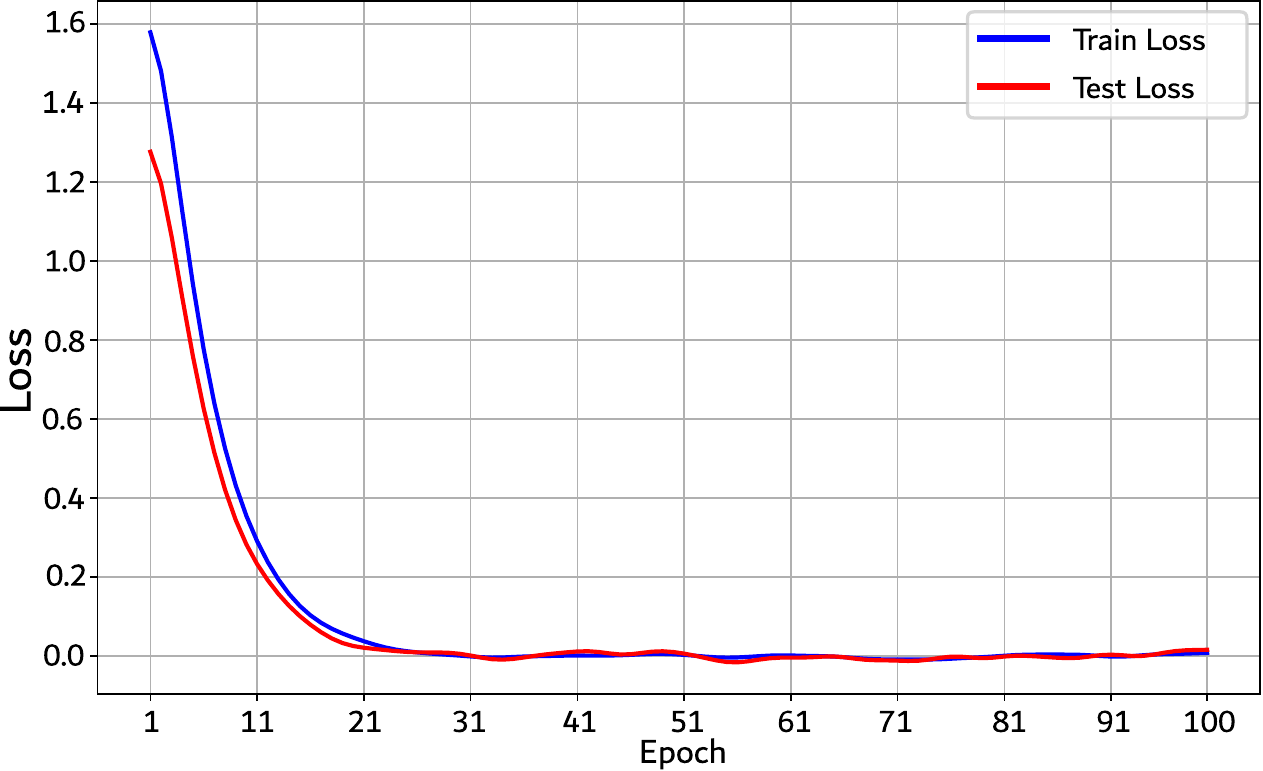}
    \caption{Loss convergence on the CPSC2018 dataset}
  \end{subfigure}
  \caption{\justifying Convergence analysis of NAT model showing training and test loss curves over 100 epochs for (a) PTB-XL dataset and (b) CPSC2018 dataset}
  \label{fig:loss_convergency}
\end{figure}

Examining the computational complexity analysis shows that our ECG-NAT model, at the fine-tuning stage based on the NAT-Tiny architecture, is more efficient than the ViT and Swin baselines under our reported settings. Figure~\ref{fig:complexity} shows that NAT-Tiny requires 29.31 million parameters, which is about 23\% fewer than ViT and 48\% fewer than Swin. It also uses 1.10GB of memory, corresponding to about 39\% and 52\% lower memory usage than ViT and Swin, respectively. NAT-Tiny requires 1.57G FLOPs, which is about 35\% lower than ViT and 50\% lower than Swin. These reductions indicate that NAT-Tiny can be a practical option in resource-constrained settings.
We also analyzed the convergence behavior during the fine-tuning stage, as shown in Figure~\ref{fig:loss_convergency}. On both the PTB-XL and CPSC-2018 datasets, the model reaches its best observed performance around 25 epochs. The training and test loss curves decrease over training and do not show clear divergence in the reported runs. This relatively fast convergence suggests that fewer epochs may be sufficient to reach strong performance. For practical ECG analysis where processing capacity is constrained, the reduced training time and resource requirements can be beneficial.
\section{Discussion and Conclusion} 
\label{se:Conclusion}
This paper proposes ECG-NAT, a novel framework integrating Neighborhood Attention Transformer with a two-stage self-supervised learning method made for classifying multi-lead ECG signals. Our approach combines local neighborhood attention with masked autoencoder pretraining on unlabeled datasets, followed by a dual-loss fine-tuning strategy that balances contrastive and cross-entropy loss to enhance feature discriminability and classification performance.
ECG-NAT addresses key challenges in ECG analysis by capturing multi-scale temporal 
features through hierarchical local attention. The architecture efficiently learns fine-grained morphological patterns within individual beats (e.g., QRS morphology, 
ST segments) while progressively expanding its receptive field across layers to capture broader rhythm-level patterns across multiple cardiac cycles, enabling accurate detection of both localized beat abnormalities and sustained rhythm irregularities. This multi-scale approach achieves computational efficiency with $O(nk)$ complexity compared to $O(n^2)$ for full self-attention while maintaining 
superior classification performance. Our real-world results demonstrate significant enhancements in several evaluation metrics. The model is more effective than other state-of-the-art models on PTB-XL, with 90.2\% accuracy and 90.4\% F1-score, and 98.6\% AUROC on CPSC2018. This model is also more computationally efficient than regular transformer architectures. Notably, ECG-NAT is very strong in low-resource settings, maintaining high performance (88.1\% accuracy) even when trained on only 1\% of labeled data. This demonstrates its applicability in clinical settings where resources are limited.

While our ECG-NAT model is highly accurate, this study did not explore its interpretability and explainability, both important considerations for clinical application. Although MAE and NAT's attention mechanisms offer a degree of inherent interpretability, there is still room to enhance the model’s overall transparency. Future work should explore structured, low-rank, or pattern-guided attention, along with disentangled latent representations aligned with interpretable ECG features. Additionally, integrating multimodal data, such as PPG, EMR, or respiratory waveforms, could enhance diagnostic performance and contextual understanding. Finally, we plan to extend our neighborhood attention mechanism to broader time-series classification tasks in healthcare and biomedical signal analysis, including EEG, EMG, PPG, and respiratory monitoring.


\section*{Acknowledgment}
Amjad Seyedi acknowledges the support by the European Union (ERC consolidator, eLinoR, no 101085607).

\bibliographystyle{unsrtnat}
\bibliography{Ref}

\end{document}